\documentclass[12pt]{article}

\usepackage{graphicx,psfrag,epsf}
\usepackage{enumerate}
\usepackage{natbib}
\usepackage{mathtools}
\usepackage{booktabs}
\usepackage{color}
\usepackage[english]{babel} % English language/hyphenation
\usepackage[protrusion=true,expansion=true]{microtype} % Better typography
\usepackage{amsmath,amsfonts,amsthm}
\usepackage{dsfont}
\usepackage{amssymb}
\usepackage{chngcntr}
\usepackage[toc,page]{appendix}
\usepackage{textcomp}
\usepackage{url}
\usepackage{hyperref}

\usepackage{array}
\usepackage{booktabs} % Horizontal rules in tables
\usepackage{setspace}
\usepackage{amsmath}
\usepackage{xcolor}
\usepackage{soul}
\usepackage{multirow}
\usepackage{enumitem}
\usepackage{microtype} % Slightly tweak font spacing for aesthetics
\usepackage[font = small,labelfont=bf,textfont=it]{subcaption}
\usepackage{xcolor}
\usepackage{tabularx}
\usepackage[font = small,labelfont=bf,textfont=it]{caption} % Custom captions under/above floats in tables or figures
\usepackage{footnote}
\usepackage{algorithm}% http://ctan.org/pkg/algorithms
\usepackage[noend]{algpseudocode}% http://ctan.org/pkg/algorithmicx
\usepackage{etoolbox}
\usepackage{tabularx}
\usepackage{authblk}
\usepackage{caption}
\usepackage{indentfirst}
\usepackage{enumitem}
\usepackage{textcomp}
\usepackage{graphicx}
\usepackage{caption}
\usepackage{subcaption}
\newlist{steps}{enumerate}{1}
\setlist[steps, 1]{label = Step \arabic*:}

\algblock{ParFor}{EndParFor}
\algnewcommand\algorithmicparfor{\textbf{for}}
\algnewcommand\algorithmicpardo{\textbf{do\ parallel}}
\algnewcommand\algorithmicendparfor{\textbf{end\ parallel\ for}}
\algrenewtext{ParFor}[1]{\algorithmicparfor\ #1\ \algorithmicpardo}
\algrenewtext{EndParFor}{\algorithmicendparfor}

\makeatletter
\def\BState{\State\hskip-\ALG@thistlm}
\DeclareMathOperator*{\argmax}{\arg\!\max}
\newcommand{\distas}[1]{\mathbin{\overset{#1}{\kern\z@\sim}}}%
\newcommand{\bm}[1]{\mathbf{#1}}

\newsavebox{\mybox}\newsavebox{\mysim}
\newcommand{\distras}[1]{%
  \savebox{\mybox}{\hbox{\kern3pt$\scriptstyle#1$\kern3pt}}%
  \savebox{\mysim}{\hbox{$\sim$}}%
  \mathbin{\overset{#1}{\kern\z@\resizebox{\wd\mybox}{\ht\mysim}{$\sim$}}}%
}
\newtheorem{theorem}{Theorem}

\newtheorem{proposition}[theorem]{Proposition}

\newcommand{\be}{\begin{equation}}
\newcommand{\ee}{\end{equation}}
    \newcommand{\bi}{\begin{itemize}}
\newcommand{\ei}{\end{itemize}}
\newcommand{\ben}{\begin{enumerate}}
\newcommand{\een}{\end{enumerate}}

\newcolumntype{K}[1]{\geq {\centering\arraybackslash}p{#1}}
\allowdisplaybreaks
\DeclareMathOperator*{\argmin}{\arg\!\min}
\makeatother

\let\oldbibliography\thebibliography
\renewcommand{\thebibliography}[1]{\oldbibliography{#1}
\setlength{\itemsep}{0pt}} %Reducing spacing in the bibliography.

\newcommand{\blind}{1}

\patchcmd{\footnotemark}{\stepcounter{footnote}}{\refstepcounter{footnote}}{}{}
\setcitestyle{square}

\begin{document}

\def\spacingset#1{\renewcommand{\baselinestretch}%
{#1}\small\normalsize} \spacingset{1}

\if1\blind
{
  \title{\bf Targeted Variance Reduction: Robust Bayesian Optimization of Black-Box Simulators with Noise Parameters}
  \small
   \author{John J. Miller\footnote{Department of Statistical Science, Duke University}\;, Simon Mak$^*$\footnote{Work supported by NSF CSSI Frameworks 2004571, NSF DMS 2210729, 2316012 and U.S. Department of Energy Grant DE-SC0024477.}
   }
  \maketitle
} \fi

\if0\blind
{
  \bigskip
  \bigskip
  \bigskip
  \begin{center}
    {\LARGE\bf Targeted Variance Reduction: Robust Bayesian Optimization of Black-Box Simulators with Noise Parameters}
\end{center}

  \medskip
} \fi

\begin{abstract}
The optimization of a black-box simulator over control parameters $\mathbf{x}$ arises in a myriad of scientific applications. In such applications, the simulator often takes the form $f(\bm{x},\boldsymbol{\theta})$, where $\boldsymbol{\theta}$ are parameters that are uncertain in practice. Robust optimization aims to optimize the objective $\mathbb{E}[f(\bm{x},\boldsymbol{\Theta})]$, where $\boldsymbol{\Theta} \sim \mathcal{P}$ is a  random variable that models uncertainty on $\boldsymbol{\theta}$. For this, existing black-box methods typically employ a two-stage approach for selecting the next point $(\mathbf{x},\boldsymbol{\theta})$, where $\mathbf{x}$ and $\boldsymbol{\theta}$ are optimized separately via different acquisition functions. As such, these approaches do not employ a joint acquisition over $(\mathbf{x},\boldsymbol{\theta})$, and thus may fail to fully exploit control-to-noise interactions for effective robust optimization. To address this, we propose a new Bayesian optimization method called Targeted Variance Reduction (TVR). The TVR leverages a novel joint acquisition function over $(\mathbf{x},\boldsymbol{\theta})$, which targets variance reduction on the objective within the desired region of improvement. Under a Gaussian process surrogate on $f$, the TVR acquisition can be evaluated in closed form, and reveals an insightful exploration-exploitation-precision trade-off for robust black-box optimization. The TVR can further accommodate a broad class of non-Gaussian distributions on $\mathcal{P}$ via a careful integration of normalizing flows. We demonstrate the improved performance of TVR over the state-of-the-art in a suite of numerical experiments and an application to the robust design of automobile brake discs under operational uncertainty.
\end{abstract}

\vfill

\newpage
\spacingset{1.55} % DON'T change the spacing!

\section{Introduction} \label{sec:intro}

Scientific computing is progressing at a remarkably rapid pace. With recent progress in scientific modeling and computing architecture, highly complex phenomena such as universe expansions \citep{kaufman2011efficient}, particle collisions \citep{ji2023graphical} and human organs \citep{chen2021function}, can now be reliably simulated via virtual experiments. A key advantage of such ``computer experiments'' \citep{gramacy2020surrogates} over traditional physical experiments is that they can \textit{control} for uncertain parameters that may be uncontrollable (e.g., environmental factors) or unknown (e.g., calibration parameters) in practice. Computer experiments, however, typically incur a high computational cost: each simulation run may take thousands of CPU hours to perform \citep{yeh2018common}. Furthermore, for such applications, decision-making typically involves optimizing the simulated response surface over a large parameter space, which requires performing many expensive simulation runs at different parameters \citep{gonzález2015Bayesian}. This computational bottleneck can thus hamper the use of computer experiments for timely scientific decision-making and discovery.

Bayesian optimization \citep{frazier2018tutorial} offers an appealing solution, and typically requires two ingredients. The first is a probabilistic ``surrogate'' model \citep{gramacy2020surrogates}, which uses limited simulation training data to fit an efficient predictive model for the expensive response surface $f(\cdot)$. The second is an acquisition function that leverages the trained surrogate model to guide the selection of subsequent query points on $f$. For surrogate model, a popular choice is the Gaussian process (GP; \citealp{RasmussenW06}), which facilitates closed-form posterior predictive equations conditional on training data. For the acquisition function, an early choice is Probability of Improvement acquisition \citep{Kushner1964}, which was subceded by the well-known Expected Improvement (EI) acquisition \citep{Jones1998}. A key reason for the popularity of EI is its closed-form expression under a GP, which permits effective optimization of subsequent query points via gradient-based methods \citep{nocedal1999numerical} without need for Monte Carlo approximations. This expression also naturally embeds the desired exploration-exploitation trade-off fundamental to reinforcement learning \citep{kearns2002near}; more on this later. Another promising acquisition function is the knowledge gradient \citep{knowledgegradient2008}, which provides a one-step optimal policy for improving the maximum of the posterior mean for $f$. Its acquisition function, however, requires costly Monte Carlo approximations, which can make the optimization of subsequent evaluation points highly challenging. This bottleneck can be alleviated to an extent via recent work \citep{gramacy2022triangulation} on a careful choice of Delaunay triangulation candidates.

% Bayesian optimization has also been extended to high dimensional \citep{Eriksson_TURBO,Nayebi19_embed} and multi-fidelity \citep{kandasamy_multifidelity,wu_multifidelity,song_multifidelity} settings.

Recall that one appeal of computer experiments is its ability to control for factors $\boldsymbol{\theta}$ that may be uncertain in reality. To make this concrete, suppose the simulator $f(\cdot)$ has two types of parameters: control parameters $\mathbf{x} \in \mathcal{X}\subset \mathbb{R}^d$ (which are controllable and to be optimized), and uncertain parameters $\boldsymbol{\theta} \in \mathcal{S}_{\Theta} \subset \mathbb{R}^q$. Here, $\boldsymbol{\theta}$ may include uncontrollable environmental factors, e.g., air humidity, and/or unknown calibration parameters, e.g., friction coefficients. In both cases, one way to account for such uncertainty is to model $\Theta$ as a ``noisy'' random variable following a carefully elicited distribution $\mathcal{P}$. For example, for air humidity, $\mathcal{P}$ might be specified via historical weather data, and for friction coefficients, $\mathcal{P}$ might be elicited via Bayesian calibration from field data. With $\mathcal{P}$ specified, a reasonable optimization problem may be:
\begin{equation}
\mathbf{x}^* = \argmax_{\mathbf{x}\in \mathcal{X}}g(\mathbf{x}) := \argmax_{\mathbf{x}\in \mathcal{X}}\mathbb{E}_{\boldsymbol{\Theta} \sim \mathcal{P}}[f(\mathbf{x},\boldsymbol{\Theta})] = \argmax_{\mathbf{x}\in \mathcal{X}} \int f(\mathbf{x},\boldsymbol{\theta}) \; d\mathcal{P}(\boldsymbol{\theta}).
\label{eq:form}
\end{equation}
This formulation is known as \textit{robust optimization} in operations research \citep{ben2009robust}. It is known \citep{taguchi1986introduction} that interactions between control and noise parameters $\mathbf{x}$ and $\boldsymbol{\theta}$ in $f$ are critical for effective robust optimization; we return to this later.

Despite its importance, there has been less work on tackling the robust optimization problem \eqref{eq:form} in the setting where the simulator $f$ is expensive. One key challenge is that the desired objective $g$ cannot be directly observed via evaluations on $f(\mathbf{x},\boldsymbol{\theta})$, as it requires averaging over $\boldsymbol{\Theta} \sim \mathcal{P}$. Given simulated data $\{f(\mathbf{x}_i, \boldsymbol{\theta}_i)\}_{i=1}^n$, the goal then is to carefully select a subsequent evaluation point $(\mathbf{x}_{n+1}, \boldsymbol{\theta}_{n+1})$ on $f$, which allows for maximization of the objective $g(\mathbf{x})$ with \textit{precision}. An early work on this is \cite{Williams2000}, who proposed a two-stage approach for optimizing $(\mathbf{x}_{n+1}, \boldsymbol{\theta}_{n+1})$. First, the control parameters $\mathbf{x}_{n+1}$ are selected to maximize the expected improvement acquisition function for $g(\mathbf{x})$, which depends on only $\mathbf{x}$. Next, with optimized $\mathbf{x}_{n+1}$, the ``noise'' parameters $\boldsymbol{\theta}_{n+1}$ are selected to minimize the prediction error on $f(\cdot)$ with $\mathbf{x}$ fixed at $\mathbf{x}_{n+1}$. Extensions of this have been explored in \cite{Groot2010BayesianMC} and \cite{swersky2013}. An important limitation of such two-stage procedures is that they do not make use of a \textit{joint} acquisition function over both $\mathbf{x}$ and $\boldsymbol{\theta}$. As such, these methods might not fully leverage the fitted interactions in $f$ between control and noise parameters, which are critical for effective robust optimization \citep{taguchi1986introduction}. This may then lead to a suboptimal choice of the next point $(\mathbf{x}_{n+1}, \boldsymbol{\theta}_{n+1})$, particularly in the presence of significant control-to-noise interactions; we will see this later in experiments. More recently, there has been work \citep{frazier-TP-2022} on extending the knowledge gradient approach \citep{knowledgegradient2008} for robust optimization, allowing for the joint selection of $(\mathbf{x}_{n+1}, \boldsymbol{\theta}_{n+1})$. Such an approach, however, does not admit a closed-form expression for the acquisition function, which introduces challenges for acquisition optimization and obfuscates interpretability, as we shall see later. 

% \cite{Pearce2017} proposes extensions of expected improvement and knowledge gradient which also allow for this joint selection, relying on sample average approximations to account for uncertainty in $\theta$ and approximate the mean objective $g$. Such approximations may under-estimate uncertainty in $\boldsymbol{\theta}$, especially as $\boldsymbol{\theta}$'s distribution increases in dimension, leading to suboptimal performance.   

To address these issues, we propose a new Bayesian optimization approach, called Targeted Variance Reduction (TVR), for tackling the robust black-box optimization problem \eqref{eq:form}. The TVR makes use of a novel acquisition function that jointly optimizes the next evaluation point $(\mathbf{x}_{n+1}, \boldsymbol{\theta}_{n+1})$, via the targeting of variance reduction \citep{gramacy2015local} on the objective $g$ within the desired region of improvement. This \textit{joint} acquisition function over both control and noise parameters can better leverage the fitted control-to-noise interactions in $f$ for guiding robust optimization. A key appeal of the TVR acquisition is that it can be evaluated in closed-form, thus enabling effective optimization of subsequent points via gradient-based methods and automatic differentiation \citep{baydin2018automatic}. This criterion further reveals a novel ``exploration-exploitation-precision'' trade-off for robust black-box optimization: it favors points $(\mathbf{x}_{n+1}, \boldsymbol{\theta}_{n+1})$ that \textit{explore} the parameter space, that \textit{exploit} promising control parameters that maximize the fitted model on objective $g$, and that improve the posterior \textit{precision} of $g$ at promising solutions. We then demonstrate the improved performance of TVR over the state-of-the-art in a suite of numerical experiments and an application to the robust design of automobile brake discs under operational uncertainty.

% (note that a function $f(\mathbf{x})$ with input perturbation can be reparametrized as $f(\mathbf{x},\boldsymbol{\theta})$, where $\boldsymbol{\theta}$ encodes perturbations of $\mathbf{x}$)

While we investigate in this work the robust optimization formulation \eqref{eq:form}, we note that there is a complementary body of work that employs alternate formulations for factoring in uncertainty on $\boldsymbol{\theta}$ for black-box optimization. This includes \cite{Marzat2013WorstcaseGO}, which optimizes $f$ under the \textit{worst-case} realization of $\boldsymbol{\theta}$ via a minimax formulation. Recent works \citep{Bogunovic2018,christianson2023robust} make use of an adversarially robust formulation, where input perturbations are set adversarially. \cite{Nogeuira2016, oliveira2019Bayesian} instead consider the setting where control inputs are randomly perturbed. To contrast, we focus on problems where uncertainty on $\boldsymbol{\theta}$ is neither worst-case nor adversarial, but instead captured by a carefully-elicited distribution $\mathcal{P}$. For such problems, the robust optimization formulation \eqref{eq:form} investigated here may be more appropriate. 

% consider optimization in the setting where the control inputs are \textit{randomly} perturbed before evaluation of $f$. Expected value optimization under random perturbations on $\mathbf{x}$ can be viewed as a special case of \eqref{eq:form}, using the reparametrization discussed above.

% , elicited from historical data for uncontrollable factors, or from Bayesian calibration for calibration factors

% \cmtS{contrast \cite{Pearce2017,oliveira2019Bayesian} - how are these different from TVR? different formulation? different methods but same formulation?} \cmtJ{Responded. Discussion of Pearce and Branke 2017 moved above, since they address the same problem as us through extending EI and KG using SAA.}

The paper is organized as follows. Section \ref{sec:backg} reviews GP modeling, its use within existing methods, and the limitations of such methods for robust black-box optimization. Section \ref{sec:tvr} presents our TVR method, including its closed-form joint acquisition function, its interpretation in terms of the exploration-exploitation-precision trade-off, and its connection to robust parameter design. Section \ref{sec:prac} provides a detailed algorithm and discusses important considerations for practical implementation. Section \ref{sec:num} demonstrates the effectiveness of TVR in a suite of numerical simulations. Section \ref{sec:app} investigates an application for robust design of automobile brake discs. Section \ref{sec:conc} concludes the paper.

\section{Background \& Motivation}
\label{sec:backg}

\subsection{Gaussian process modeling}
We first provide a brief review of Gaussian process surrogate modeling. Let $f(\mathbf{x},\boldsymbol{\theta})$ denote the scalar output of the expensive computer code, where $\mathbf{x} \in \mathcal{X} \subset \mathbb{R}^d$ are control parameters and $\boldsymbol{\theta} \in \mathcal{S}_\Theta \subset \mathbb{R}^q$ represent parameters uncertain in practice. As $f(\cdot)$ is black-box, we adopt the Gaussian process prior $f(\cdot) \sim \textup{GP}\{\mu,k(\cdot,\cdot)\}$; see \cite{gramacy2020surrogates} for details. Here, $\mu$ is a mean parameter for the GP, and $k(\cdot,\cdot)$ is its covariance function over the joint parameters $(\mathbf{x},\boldsymbol{\theta})$. Suppose the simulator is run at the design points $\{(\mathbf{x}_i,\boldsymbol{\theta}_i)\}_{i=1}^n$, yielding data $\mathbf{f}_n = [f(\mathbf{x}_i,\boldsymbol{\theta}_i)]_{i=1}^n$. Conditional on such data, the posterior predictive distribution of $f$ at a new point $(\mathbf{x}_{\rm new},\boldsymbol{\theta}_{\rm new})$ can be shown to be:
\begin{equation}
f(\mathbf{x}_{\rm new},\boldsymbol{\theta}_{\rm new})| \mathbf{f}_n \sim \mathcal{N}\{\tilde{\mu}_n(\mathbf{x}_{\rm new},\boldsymbol{\theta}_{\rm new}), \tilde{s}^2_n(\mathbf{x}_{\rm new},\boldsymbol{\theta}_{\rm new})\}.
\label{eq:postgp}
\end{equation}
Here, the posterior mean and variance take the closed-form expressions:
\begin{align}
\begin{split}
\tilde{\mu}_n(\mathbf{x}_{\rm new},\boldsymbol{\theta}_{\rm new}) &= \mu + \mathbf{k}_n^T(\mathbf{x}_{\rm new},\boldsymbol{\theta}_{\rm new}) \mathbf{K}_n^{-1}(\mathbf{f}_n-\mu \mathbf{1}),\\
\tilde{s}^2_n(\mathbf{x}_{\rm new},\boldsymbol{\theta}_{\rm new}) &= k((\mathbf{x}_{\rm new},\boldsymbol{\theta}_{\rm new}),(\mathbf{x}_{\rm new},\boldsymbol{\theta}_{\rm new})) - \mathbf{k}_n^T(\mathbf{x}_{\rm new},\boldsymbol{\theta}_{\rm new}) \mathbf{K}_n^{-1}\mathbf{k}_n(\mathbf{x}_{\rm new},\boldsymbol{\theta}_{\rm new}),
\end{split}
\end{align}
where $\mathbf{K}_n = [k((\mathbf{x}_i,\boldsymbol{\theta}_i),(\mathbf{x}_j,\boldsymbol{\theta}_j))]_{i,j=1}^n$ and $\mathbf{k}_n(\mathbf{x}_{\rm new},\boldsymbol{\theta}_{\rm new}) = [k((\mathbf{x}_i,\boldsymbol{\theta}_i),(\mathbf{x}_{\rm new},\boldsymbol{\theta}_{\rm new}))]_{i=1}^n$.

For the robust optimization problem \eqref{eq:form}, a key challenge is that the desired objective $g(\mathbf{x}) = \mathbb{E}_{\boldsymbol{\Theta} \sim \mathcal{P}}[f(\mathbf{x},\boldsymbol{\theta})]$ cannot be directly observed from evaluations on the simulator $f$. Consider the specific setting where $\mathcal{P}$ is a \textit{discrete} probability measure with finite support $\{\boldsymbol{\theta}_m\}_{m=1}^M$ and probability masses $\{p_m\}_{m=1}^M$. In this case, one can show that the posterior predictive distribution of the objective $g(\mathbf{x}_{\rm new})$ given data $\mathbf{f}_n$ takes the form:
\begin{equation}
g(\mathbf{x}_{\rm new})| \mathbf{f}_n \sim \mathcal{N}\{{\mu}_n(\mathbf{x}_{\rm new}), {s}^2_n(\mathbf{x}_{\rm new})\},
\label{eq:postg}
\end{equation}
where the posterior mean and variance of $g(\mathbf{x}_{\rm new})$ admit the closed-form expressions:
\small
\begin{align}
\begin{split}
    \mu_n(\mathbf{x}_{\rm new}) &= \sum_{m=1}^M p_m \tilde{\mu}_n(\mathbf{x}_{\rm new},\boldsymbol{\theta}_m),\\
    s^2_n(\mathbf{x}_{\rm new}) &= \sum_{m=1}^M \sum_{m'=1}^M p_m p_{m'} \left\{ k((\mathbf{x}_{\rm new},\boldsymbol{\theta}_m),(\mathbf{x}_{\rm new},\boldsymbol{\theta}_{m'})) - \mathbf{k}_n^T(\mathbf{x}_{\rm new},\boldsymbol{\theta}_m) \mathbf{K}_n^{-1}\mathbf{k}_n(\mathbf{x}_{\rm new},\boldsymbol{\theta}_{m'}) \right\}.
    \label{eq:postgcl}
\end{split}
\end{align}
\normalsize
Thus, when $\boldsymbol{\Theta}$ is a finite discrete random variable, Equations \eqref{eq:postg} and \eqref{eq:postgcl} permit closed-form prediction and uncertainty quantification of objective $g$ from simulator evaluations $\mathbf{f}$.

In practice, however, $\boldsymbol{\Theta}$ often consists of continuous parameters, coupled with a potentially complex distribution $\mathcal{P}$, e.g., arising from a Bayesian calibration of the simulation system. We will extend in Section \ref{sec:tvrgp} the above closed-form predictive approach for this more complex setting of continuous $\boldsymbol{\Theta}$ with complex distributions.

\subsection{Existing state-of-the-art}
\label{sec:exist}

There is an existing body of work that tackle the black-box robust optimization problem \eqref{eq:form}. The underlying idea is to leverage a GP surrogate on the black-box response surface $f(\cdot)$ for guiding the selection of the next evaluation point $(\mathbf{x}_{n+1},\boldsymbol{\theta}_{n+1})$. An early work is \cite{Williams2000}, who proposed a two-stage approach for selecting $(\mathbf{x}_{n+1},\boldsymbol{\theta}_{n+1})$ when $\boldsymbol{\Theta}$ is a finite discrete random variable. In the first stage, the control parameters $\mathbf{x}_{n+1}$ are chosen to maximize the expected improvement acquisition on $g$, i.e.:
\begin{equation}
\mathbf{x}_{n+1} \leftarrow \argmax_{\mathbf{x} \in \mathcal{X}} \mathbb{E}[\max\{0,g(\mathbf{x})-g_n^*\}|\mathbf{f}_n],
\label{eq:stg1}
\end{equation}
where $g^*_n= \max\{g(\mathbf{x}_1),\dots,g(\mathbf{x}_n)\}$. Since $g$ is not directly observed, a Monte Carlo approximation is employed for \eqref{eq:stg1} using the predictive distribution \eqref{eq:postg}. In the second stage, with selected control parameters $\mathbf{x}_{n+1}$, the noise parameters $\boldsymbol{\theta}_{n+1}$ are chosen to minimize:
\begin{equation}
\boldsymbol{\theta}_{n+1} \leftarrow \argmin_{\boldsymbol{\theta}\in\boldsymbol{\Theta}} \mathbb{E}[\{\mu_{n+1}(\mathbf{x}_{n+1}) - g(\mathbf{x}_{n+1})\}^2|\mathbf{f}_n].
\label{eq:stg2}
\end{equation}
This choice of $\boldsymbol{\theta}$ targets reduction of prediction error for the surrogate model on $g$ along the hyperplane $\mathbf{x} = \mathbf{x}_{n+1}$. One then evaluates the simulator at the selected point $(\mathbf{x}_{n+1},\boldsymbol{\theta}_{n+1})$, and this two-stage procedure is repeated until the computational budget is exhausted. \cite{Groot2010BayesianMC} explore extensions of this two-stage procedure for broader distributions using a generalized EI criterion, albeit without closed-form acquisition functions at each stage.

A limitation of such two-stage approaches, however, is the absence of a \textit{joint} acquisition function over both control parameters $\mathbf{x}$ and noise parameters $\boldsymbol{\theta}$. This introduces two key disadvantages when selecting the next point $(\mathbf{x}_{n+1},\boldsymbol{\theta}_{n+1})$. First, this may cause instability in the iterative optimization steps \eqref{eq:stg1} and \eqref{eq:stg2}, since neither step targets a \textit{common} joint acquisition over $(\mathbf{x},\boldsymbol{\theta})$. Second, in separating the optimization of $\mathbf{x}_{n+1}$ and $\boldsymbol{\theta}_{n+1}$, two-stage approaches may not fully leverage the underlying \textit{interactions} between control and noise parameters in $f$. Such interactions are critical for achieving effective robust parameter design \citep{taguchi1986introduction}, a highly related problem that we discuss further in Section \ref{sec:rob}. The lack of a joint acquisition function can thus lead to a suboptimal choice of the next evaluation point $(\mathbf{x}_{n+1},\boldsymbol{\theta}_{n+1})$, particularly in the presence of large control-to-noise interactions. We will explore this in a later motivating illustration.

Another recent work \citep{frazier-TP-2022} investigates an alternate strategy by extending the knowledge gradient approach in \cite{knowledgegradient2008}. Here, the next query point maximizes the following:
\begin{equation}
(\mathbf{x}_{n+1},\boldsymbol{\theta}_{n+1}) \leftarrow \argmax_{(\mathbf{x},\boldsymbol{\theta})\in \mathcal{X}\times\mathcal{S}_{{\Theta}}} \mathbb{E}\left[\max_{\mathbf{x}'\in\mathcal{X}}\mu_{n+1}(\mathbf{x}') -\max_{\mathbf{x}'\in\mathcal{X}}\mu_{n}(\mathbf{x}') \Big| \mathbf{f}_n,f(\mathbf{x},\boldsymbol{\theta}) \right].
\label{eq:BQO}
\end{equation}
In words, the next point $(\mathbf{x}_{n+1},\boldsymbol{\theta}_{n+1})$ maximizes improvement in the surrogate optimum given a new evaluation on $f$. One disadvantage of such an approach is that the acquisition in \eqref{eq:BQO} cannot be evaluated in closed form and requires Monte Carlo approximation. This introduces two limitations. First, its optimization for sequential queries becomes more challenging as each acquisition evaluation can be expensive; this may in turn result in lower quality query points, as we shall see later. Second, the lack of a closed-form acquisition may obfuscate \textit{interpretability}. A primary reason for the popularity of the expected improvement method \citep{Jones1998} is its natural interpretation in terms of the exploration-exploitation trade-off via its closed-form acquisition \citep{chen2023}. We shall address this later via a new closed-form and interpretable acquisition function for robust black-box optimization.

\subsection{A motivating illustration}\label{sec:motivation}

\begin{figure}[!t]
    \centering
    \includegraphics[width=0.80\textwidth]{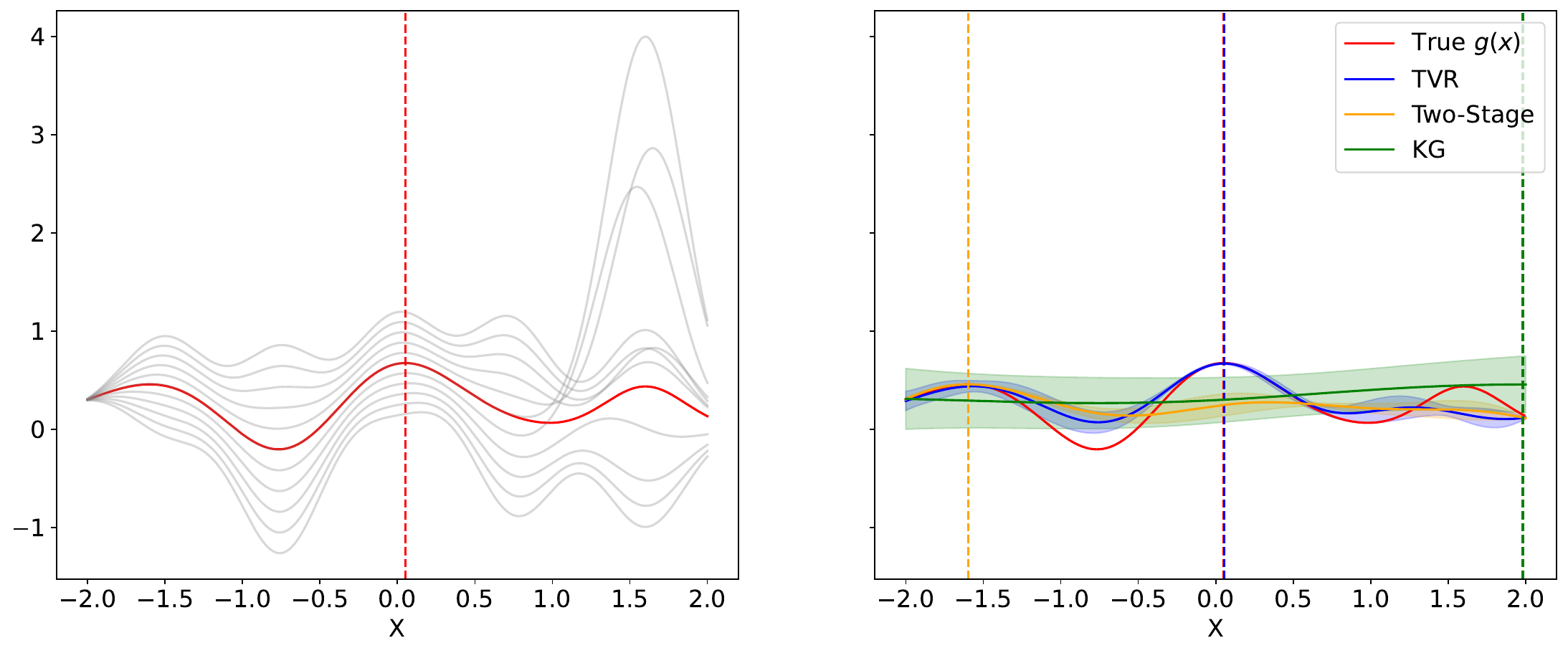}
    \caption{\textup{[Left]} Realizations of the test function $f(x,{\Theta})$ in \eqref{eq:mot} for different samples of $\Theta$. Solid red curves mark the desired objective $g(x) = \mathbb{E}[f(x ,\Theta)]$, and the dotted line shows the optimum $x^* = 0.05$. \textup{[Right]} Visualizing the fitted GP model on $g(x)$ and its corresponding chosen solution $x_n^*$ for various compared methods. Here, solid curves mark the posterior mean $\mu_n(x)$, the shaded regions mark its 95\% confidence region, and the dotted lines show the chosen solution $x_n^*$.}
    % Optimization performance for the implied $g(x)$ after $T=25$ evaluations of $f$ (including an initial design of $10$ observations). The naive application of Bayesian optimization with the Knowledge Gradient \citep{knowledgegradient2008} acquisition function and the two-stage acquisition function (which accounts for problem structure) both fail to find the global maximum, even after 25 evaluations of $f$. By accounting for $\theta$'s distribution and jointly selecting $(\mathbf{x}_{n+1},\boldsymbol{\theta}_{n+1})$ utilizing the proposed TVR acquisition function, a high-quality solution is found.
    \label{fig:trig_TVR_demo}
\end{figure}

To highlight the importance of control-to-noise interactions for robust optimization, we investigate the following test function: 
\small
\begin{align}
\begin{split}
f(x,\theta) & = \frac{4}{{\left( {\theta^4}/{2} + 1 \right)}} \exp \left\{ -8 \left( x + \frac{\theta}{20} - \frac{8}{5} \right)^2 \right\} + \frac{1}{2}\exp\left\{-2\left(x + \frac{\theta}{50} + \frac{3}{2}\right)^2\right\} \\
& \quad  + \frac{5}{7}\exp(-3x^2) - \frac{1}{2}\exp\left\{-4\left(x+\frac{3}{4}\right)^2\right\}\\
& \quad - \frac{\theta}{5} \Bigg[\frac{1}{2} \exp\left\{-8\left(x+\frac{3}{2}\right)^2\right\} + \frac{1}{2}\exp(-8x^2) +\exp\left\{-8\left(x-\frac{3}{4}\right)^2\right\}  \\
& \quad \quad \quad \quad + \exp\left\{-8\left(x+\frac{3}{4}\right)^2\right\} + \exp\left\{-8\left(x-\frac{8}{5}\right)^2\right\} \Bigg].
\label{eq:mot}
\end{split}
\end{align}
\normalsize
Here, there is a single control and noise parameter, with significant control-to-noise interactions by construction. Figure \ref{fig:trig_TVR_demo} (left) visualizes such interactions: at different $\theta$, the function slices $f(\cdot,\theta)$ can vary greatly over $x$, which results in considerably different maximizers in $x$ for each slice. We set the feasible region as $\mathcal{X} = [-2,2]$, and adopt a discrete distribution on $\Theta$, with $\mathbb{P}(\Theta = m) \propto |m|+1$, $m = -5, 4, \cdots, 4, 5$. We then test two methods: (i) the two-stage approach from \cite{Williams2000}, and (ii) a ``naive'' noisy knowledge gradient approach, which treats observations on $f(x,\theta)$ as noisy evaluations on some function $\tilde{f}(x)$ depending on only $x$. The latter is thus an extreme case where all control-to-noise interactions are ignored. (Broader comparisons with other methods are provided in Section \ref{sec:num}; here, we focus on showing the importance of control-to-noise interactions.) All methods begin with an initial equally-spaced design of 10 runs, then proceed for 25 sequential runs.

% One approach to optimize a function taking noise parameters is to simply ignore the structured influence of $\boldsymbol{\theta}$ on $f$ by treating $f(\mathbf{x},\boldsymbol{\theta})$ as a noisy function of only $\mathbf{x}$, where observed values are of the form $y(\mathbf{x}) = f(\mathbf{x})+\epsilon$ with $\mathbb{E}[\epsilon]=0$. Within this framework, one can apply any of the tools from the Bayesian Optimization literature that are able to accommodate noisy function evaluations. For certain problems this approach may suffice, particularly if the influence of $\boldsymbol{\theta}$ on the objective is relatively weak or if there is little nontrivial interaction between $\mathbf{x}$ and $\boldsymbol{\theta}$. 

% Another approach is to incorporate the existence of $\boldsymbol{\theta}$ and knowledge of its distribution by using a Two-stage acquisition procedure, similar to what is proposed in \cite{Williams2000,Groot2010BayesianMC,swersky2013}. One such procedure is to first select $\mathbf{x}_{n+1}$ by maximizing EI on the objective $g$ (see \eqref{eq:stg1}) and then selecting the value of $\boldsymbol{\theta}_{n+1}$ which maximizes the reduction in variance reduction at $g(\mathbf{x}_{n+1})$ while keeping $\mathbf{x}_{n+1}$ fixed. In this sort of approach, the choice of $\mathbf{x}_{n+1}$ affects the choice of $\boldsymbol{\theta}_{n+1}$ but not vice versa. A consequence of this is a reduction in the amount information available when selecting $\mathbf{x}_{n+1}$.

Figure \ref{fig:trig_TVR_demo} (right) shows, for each method, the fitted surrogate on the objective $g(x)$ given sampled points, along with the chosen solution $x_n^*$ taken as the maximizer of the surrogate mean $\mu_n(x) = \mathbb{E}[g(x)|\mathbf{f}_n]$; see \eqref{eq:postgcl}. The naive knowledge gradient approach performs poorly, which highlights the importance of leveraging control-to-noise interactions for robust optimization. The two-stage approach selects the solution $x_n^* \approx -1.6$, which is a local maximizer of $g$ (see Figure \ref{fig:trig_TVR_demo} left) but has noticeably lower objective value compared to the global maximizer at $\bm{x}^* = 0.051$. One reason for this is the absence of a \textit{joint} acquisition over $\mathbf{x}$ and $\boldsymbol{\theta}$, which may inhibit the full exploitation of control-to-noise interactions for sequential queries, and thus cause the procedure to be stuck in a local maxima. The TVR, introduced next, achieves a near-optimal solution of $x^*_n \approx 0.053$, via a new joint (and closed-form) acquisition that better leverages the underlying fitted control-to-noise interactions in $f$.

\section{Targeted Variance Reduction}
\label{sec:tvr}
We first outline a flexible surrogate modeling framework that accommodates for continuous $\boldsymbol{\Theta}$ with general distributions, then present the novel closed-form TVR acquisition function. We then investigate the exploration-exploitation-precision trade-off captured by the TVR, and highlight useful connections with robust parameter design \citep{taguchi1986introduction}.

\subsection{Modeling framework}
\label{sec:tvrgp}

Consider the general setting where $\boldsymbol{\Theta}$ is continuous with probability distribution $\mathcal{P}$. The first step is to specify a transformation $\mathcal{T}$ such that $\mathcal{T}(\mathbf{Z}) = \boldsymbol{\Theta}$, where $\mathbf{Z} \sim \mathcal{N}(\mathbf{0},\mathbf{I})$ is the standard isotropic normal distribution. When the cumulative distribution function (c.d.f) of $\mathcal{P}$ (call this $P$) is known, then a natural choice of transformation is $\mathcal{T}(\mathbf{z}) = P^{-1}\{\Phi(\mathbf{z})\}$, where $\Phi(\cdot)$ is the c.d.f of the standard normal $\mathbf{Z}$. Specifying such a transformation becomes more challenging when the c.d.f $P$ is not known. This may arise when $\boldsymbol{\theta}$ consists of parameters that are calibrated in a Bayesian fashion, in which case the normalizing constant of $P$ would typically be unknown. For this, a recent development in generative machine learning called \textit{normalizing flows} (see, e.g., \citealp{Kobyzev2021}) offers an appealing solution. Normalizing flows can learn the desired invertible transformation $\mathcal{T}$ using samples from $\mathcal{P}$, via a carefully-designed neural network architecture. Such methods enjoy strong empirical performance \citep{Kobyzev2021}, with supporting theoretical guarantees for a broad range of continuous distributions \citep{huang2018neural,jaini2019}. Given the ability to sample from $\mathcal{P}$, the transformation $\mathcal{T}$ can be trained offline via normalizing flows, prior to experimentation. With $\mathcal{T}$ in hand, we can then perform the subsequent surrogate modeling and sequential queries on the \textit{reparametrized} response surface $\tilde{f}(\mathbf{x},\mathbf{z}) = f(\mathbf{x},\mathcal{T}(\mathbf{z}))$.

The reason behind this reparametrization is to facilitate closed-form surrogate modeling of the objective $g$. As before, let us adopt a GP model on $\tilde{f}$, i.e., $\tilde{f}(\cdot) \sim \text{GP}\{\mu,k(\cdot,\cdot)\}$, where $k(\cdot,\cdot)$ is the anisotropic squared-exponential kernel widely used in computer experiments:
\begin{equation}
k((\mathbf{x},\mathbf{z}),(\mathbf{x}',\mathbf{z}')) = \sigma^2 \exp\left\{-\sum_{j=1}^d \frac{(x_j-x_j')^2}{2\ell_j^2} - \sum_{l=1}^q \frac{(z_l - z_l')^2}{2\gamma_l^2} \right\}.
\end{equation}
Suppose the simulator is run at design points $\{(\mathbf{x}_i,\boldsymbol{\theta}_i)\}_{i=1}^n$, yielding data on the reparametrized response surface $\tilde{\mathbf{f}}_n = [\tilde{f}(\mathbf{x}_i,\mathbf{z}_i)]_{i=1}^n$, where $\mathbf{z}_i = \mathcal{T}^{-1}(\boldsymbol{\theta}_i)$. Conditional on such data, one can show that the predictive distribution of $g(\mathbf{x}_{\rm new})$ is:
\begin{equation}
g(\mathbf{x}_{\rm new})| \tilde{\mathbf{f}}_n \sim \mathcal{N}\{{\mu}_n(\mathbf{x}_{\rm new}), {s}^2_n(\mathbf{x}_{\rm new})\}.
\label{eq:postg2}
\end{equation}
Here, the posterior mean and variance can be evaluated in closed form as:
\begin{align}
\begin{split}
\mu_n(\mathbf{x}_{\rm new}) &= \mu + \mathbf{h}_n^T(\mathbf{x}_{\rm new}) \mathbf{K}_n^{-1}[\mathbf{f}_n - \mu \mathbf{1}],\\
s^2_n(\mathbf{x}_{\rm new}) &= s^2_0(\mathbf{x}_{\rm new},\mathbf{x}_{\rm new}) - \mathbf{h}_n^T(\mathbf{x}_{\rm new}) \mathbf{K}_n^{-1} \mathbf{h}_n(\mathbf{x}_{\rm new}),
\label{eq:postgcl2}
\end{split}
\end{align}
where, with $\mathbf{h}_n(\mathbf{x}_{\rm new}) = [h((\mathbf{x}_i,\mathbf{z}_i),\mathbf{x}_{\rm new})]_{i=1}^n$, we have:
\small
\begin{align}
\begin{split}
h((\mathbf{x},\mathbf{z}),\mathbf{x}') &= \sigma^2 \exp\left\{-\sum_{j=1}^d\frac{(x_j-x_j')^2}{2\ell_j^2} \right\} \prod_{l=1}^q (\gamma_l^{-2}+1)^{-1/2} 
\exp\left\{-\frac{z_l^2 }{2}(\gamma_l^{-2}-(\gamma_l^2 +\gamma_l^4)^{-1})\right\},\\
s_0^2(\mathbf{x},\mathbf{x}') &= \sigma^2 \exp\left\{-\sum_{j=1}^d\frac{(x_j-x_j')^2}{2\ell_j^2} \right\}  \prod_{l=1}^q \left[(\gamma_l^{-2}+1)\left(\gamma_l^{-2}+1 - (\gamma_l^2 +\gamma_l^4)^{-1}\right) \right]^{-1/2}.
\label{eq:postgcl3}
\end{split}
\end{align}
\normalsize
The derivation of this predictive distribution can be found in the Supplementary Materials.

 % \cmtS{do we need some comments on if $\boldsymbol{\Theta}$ has mixed variables? other combinations of $k$ and $\mathcal{P}$ (e.g. $k$ matern and $\mathcal{P}$ t-distributed) that also might work?} \cmtJ{Responded below.} 
The key importance of Equations \eqref{eq:postg2}-\eqref{eq:postgcl3} is that, for a general specification of continuous $\boldsymbol{\Theta} \sim \mathcal{P}$, one can obtain closed-form predictive equations on the objective $g(\cdot)$, provided the required transformation $\mathcal{T}$. Such equations then nicely translate into an interpretable and closed-form acquisition function for TVR, as we show next. In the simpler case of discrete $\boldsymbol{\Theta}$, closed-form predictive equations follow directly from \eqref{eq:postgcl}. Such expressions also extend analogously when $\boldsymbol{\Theta}$ contains mixed variables, i.e., some variables are continuous while others are discrete.

\subsection{Acquisition function}

We present next the new TVR acquisition function, which provides a joint acquisition over $\mathbf{x}$ and $\boldsymbol{\theta}$ using the above model. Here, we assume the reparametrization from Section \ref{sec:tvrgp} has already been applied via transformation $\mathcal{T}$, thus without loss of generality, we assume in the following that the reparametrized simulator takes the form $f(\mathbf{x},\boldsymbol{\theta})$ with $\boldsymbol{\Theta} \sim \mathcal{N}(\mathbf{0},\mathbf{I})$.

Recall from Section \ref{sec:exist} that existing two-stage approaches iteratively target (i) the improvement in the objective $g$, and (ii) the precision (or accuracy) of the underlying surrogate model. The TVR combines both goals via the utility function:
\begin{equation}
U_{\rm TVR}(\mathbf{x},\boldsymbol{\theta}) = \mathds{1}_{\{g(\mathbf{x}) >  g(\mathbf{x}^*_n) \}}\text{VR}_n(\mathbf{x},\boldsymbol{\theta}),
\label{eq:util}
\end{equation}
where $\mu_n(\mathbf{x}) = \mathbb{E}[g(\mathbf{x}_{\rm new})|\mathbf{f}_n]$ is the predictor from \eqref{eq:postgcl2}, and $\mathbf{x}^*_n = \argmax_{\mathbf{x}} \mu_n(\mathbf{x})$ is the predicted solution. Here, $\text{VR}_n(\mathbf{x},\boldsymbol{\theta})$ is the \textit{variance reduction} criterion employed in \cite{gramacy2015local} on objective $g$, given by:
\begin{equation}
\text{VR}_n(\mathbf{x},\boldsymbol{\theta}) = \text{Var}[g(\mathbf{x})|\mathbf{f}_n]-\text{Var}[g(\mathbf{x})| \mathbf{f}_n, f(\mathbf{x},\boldsymbol{\theta})],
\label{eq:vr}
\end{equation}
where both variance terms in \eqref{eq:vr} admit closed-form expressions from \eqref{eq:postg2} and \eqref{eq:postgcl2}.

The utility function \eqref{eq:util} can be understood in two parts. The first term in \eqref{eq:util} provides an indicator function for the region of improvement $\{\mathbf{x}: g(\mathbf{x}) >  g(\mathbf{x}^*_n) \}$, where the objective $g$ at $\mathbf{x}$ is larger than the predicted maximum $g(\mathbf{x}_n^*)$. Provided $\mathbf{x}$ lies within this region of improvement, the second term in \eqref{eq:util} then quantifies the variance reduction in the objective $g$, given the next design point for simulator $f$ is at $(\mathbf{x},\boldsymbol{\theta})$. Intuitively, these two terms in the TVR utility \textit{jointly} capture the aforementioned two goals (i) and (ii): we wish to target control parameters $\mathbf{x}$ that lead to objective improvement (namely, goal (i)), and within this region of improvement, we wish to identify joint parameters $(\mathbf{x},\boldsymbol{\theta})$ that increase precision of the desired objective (namely, goal (ii)).

Taking the posterior expectation of the TVR utility function \eqref{eq:util} given data $\mathbf{f}_n$, we obtain the following expression for the TVR acquisition, assuming $\mathbf{x} \neq \mathbf{x}_n^*$:
\begin{equation}
\text{TVR}(\mathbf{x},\boldsymbol{\theta}) = \mathbb{E}[U_{\rm TVR}(\mathbf{x},\boldsymbol{\theta})|\mathbf{f}_n] = \text{VR}_n(\mathbf{x},\boldsymbol{\theta})\Phi\left(\frac{\mu_n(\mathbf{x})-\mu_n(\mathbf{x}_n^*)}{[s^2_n(\mathbf{x}_n^*)+s^2_n(\mathbf{x})-2s_n(\mathbf{x},\mathbf{x}_n^*)]^{1/2}} \right),
\label{eq:tvr}
\end{equation}
where $s_n(\mathbf{x},\mathbf{x}') = \text{Cov}[g(\mathbf{x}),g(\mathbf{x}')|\mathbf{f}_n] = s^2_0(\mathbf{x},\mathbf{x}') - \mathbf{h}_n^T(\mathbf{x}) \mathbf{K}_n^{-1} \mathbf{h}_n(\mathbf{x}')$. Note that the latter term in \eqref{eq:tvr} (involving $\Phi(\cdot)$) is similar to the Probability of Improvement acquisition in \citep{Kushner1964}, a predecessor of the EI; its use within the TVR admits a nice interpretation, which we discuss later. With this, the next control and noise parameters $(\mathbf{x}_{n+1},\boldsymbol{\theta}_{n+1})$ can be \textit{jointly} selected by maximizing the TVR acquisition function, i.e.:
\begin{equation}
(\mathbf{x}_{n+1},\boldsymbol{\theta}_{n+1}) \leftarrow \argmax_{(\mathbf{x},\boldsymbol{\theta})} \text{TVR}(\mathbf{x},\boldsymbol{\theta}).
\label{eq:tvrnext}
\end{equation}
A slight extension of $\text{TVR}(\mathbf{x},\boldsymbol{\theta})$ will be employed later for implementation (see Section \ref{sec:disc}).

The proposed TVR acquisition \eqref{eq:tvrnext} addresses the aforementioned limitations of existing black-box methods for robust optimization. First, from the closed-form expression \eqref{eq:tvr}, note that each term in $\text{TVR}(\mathbf{x},\boldsymbol{\theta})$ can be evaluated quickly given the fitted surrogate model on $g(\mathbf{x})$. As such, the next evaluation point $(\mathbf{x}_{n+1},\boldsymbol{\theta}_{n+1})$ can be efficiently optimized via \eqref{eq:tvrnext} using gradient-based methods and automatic differentiation \citep{baydin2018automatic}. This addresses a limitation of the earlier knowledge gradient approach in \cite{frazier-TP-2022}. Second, the TVR acquisition function is \textit{jointly} defined over both the control and noise parameters $\mathbf{x}$ and $\boldsymbol{\theta}$; this is in contrast to existing two-stage procedures \citep{Williams2000,Groot2010BayesianMC,swersky2013}, which optimize $\mathbf{x}$ and $\boldsymbol{\theta}$ \textit{separately} based on different criteria. As we show later in numerical experiments, this new joint acquisition function appears to better leverage the fitted control-to-noise interactions in $f$ for effective robust optimization.

The TVR acquisition also highlights an insightful \textit{exploration-exploitation-precision} trade-off for robust black-box optimization. From \eqref{eq:tvr}, the maximization of the acquisition $\text{TVR}(\mathbf{x},\boldsymbol{\theta})$ involves several factors. The first factor involves finding control parameters $\mathbf{x}$ for which $\mu_n(\mathbf{x}) = \mathbb{E}[g(\mathbf{x})|\mathbf{f}_n]$ (and thus the numerator term $\mu_n(\mathbf{x})-\mu_n(\mathbf{x}_n^*)$) is large. This is precisely the notion of \textit{exploitation}: we wish to find $\mathbf{x}$ that yield large fitted values for the objective $g$. Observing that $\mu_n(\mathbf{x})-\mu_n(\mathbf{x}_n^*) \leq 0$ by definition of $\mathbf{x}_n^*$, the second factor involves finding control parameters $\mathbf{x}$ for which the denominator term $s^2_n(\mathbf{x}_n^*)+s^2_n(\mathbf{x})-2s_n(\mathbf{x},\mathbf{x}_n^*)$ is large. This can be interpreted as \textit{exploration}: we wish to find $\mathbf{x}$ with greatest uncertainty in $g$, i.e., with large $s^2_n(\mathbf{x}) = \text{Var}[g(\mathbf{x})|\mathbf{f}_n]$, particularly solutions that are further away from the predicted minimizer $\mathbf{x}_n^*$, i.e., with small $s_n(\mathbf{x},\mathbf{x}_n^*)$. The final factor involves finding $(\mathbf{x},\boldsymbol{\theta})$ for which $\text{VR}_n(\mathbf{x},\boldsymbol{\theta})$ is large. This encourages objective \textit{precision}: for control parameters $\mathbf{x}$ with good exploration and exploitation, we then wish to find noise parameters $\boldsymbol{\theta}$ that decrease variance (i.e., increase precision) on the desired objective $g(\mathbf{x})$, which is not directly observed from data. This thus provides a novel extension of the well-known exploration-exploitation trade-off in reinforcement learning \citep{kearns2002near} for the robust black-box optimization setting at hand.

\subsection{Connection to robust parameter design}
\label{sec:rob}

There are also insightful connections between the current black-box robust optimization problem \eqref{eq:form} (and our solution via the TVR) and the classical problem of \textit{robust parameter design} (see, e.g., Chapter 11 of \citealp{wu2011experiments}). Robust parameter design (RPD) was popularized by \cite{taguchi1986introduction} for quality engineering, and aims to reduce sensitivity of a response variable in the presence of uncontrollable noise factors $\boldsymbol{\theta}$ via a careful specification of control parameters $\mathbf{x}$. This concept was critical in catalyzing the product quality revolution in the 1980's, particularly within the automobile and electronics industries \citep{Mori_taguchibook}. Fundamental to RPD is the presence of control-to-noise interactions, i.e., interactions in the response surface $f$ between $\mathbf{x}$ and $\boldsymbol{\theta}$. Intuitively, it is clear why these interactions are needed: without such effects, the response surface $f(\mathbf{x},\cdot)$ would behave similarly over $\boldsymbol{\theta}$ for any choice of $\mathbf{x}$, and thus there is little potential for reducing sensitivity of the response to noise $\boldsymbol{\theta}$ via a careful specification of control parameters $\mathbf{x}$.

The robust optimization problem \eqref{eq:form} can then be viewed as a specific case of robust parameter design, where one is interested in the expected response in the presence of uncontrollable and random noise $\boldsymbol{\Theta} \sim \mathcal{P}$. Given this connection, it is not surprising that the exploitation of control-to-noise interactions is similarly important for the current robust optimization problem. Existing two-stage approaches (see Section \ref{sec:exist}), which separately select $\mathbf{x}_{n+1}$ and $\boldsymbol{\theta}_{n+1}$ using different criteria, may thus fail to fully leverage the fitted control-to-noise interactions for effective robust optimization. The TVR directly addresses this need via a principled \textit{joint} acquisition function over both $\mathbf{x}$ and $\boldsymbol{\theta}$, to better exploit such interactions for guiding sequential queries from the simulator $f$. The increased flexibility of a GP surrogate model (used in the TVR) may also aid in the estimation of such control-to-noise interactions, compared to more classical response surface models used in the robust parameter design literature (see, e.g., \citealp{wu2011experiments}).

\section{Practical Implementation}
\label{sec:prac}
We now discuss several important practical considerations for effective implementation of the TVR. This includes a slight extension of the TVR to avoid potential discontinuities for optimization, a batch sampling approach, then a detailed algorithm for implementation.

\subsection{Acquisition extension and optimization}
\label{sec:disc}

Note that the TVR acquisition function \eqref{eq:tvr} technically holds only when the control parameters $\mathbf{x}$ satisfy $\mathbf{x} \neq \mathbf{x}_n^*$, where $\mathbf{x}_n^*$ is the currently chosen solution. When $\mathbf{x}$ equals $\mathbf{x}_n^*$, the TVR as defined above reduces to 0, since $\mathds{1}_{\{g(\mathbf{x}) >  g(\mathbf{x}^*_n) \}}$ would equal $0$. This introduces an axis of discontinuity for $\text{TVR}(\mathbf{x},\boldsymbol{\theta})$ that spikes down along $\mathbf{x} = \mathbf{x}_n^*$, which not only poses an issue for acquisition optimization, but is also quite unintuitive. With further consideration, however, such a discontinuity can easily be remedied for two reasons. First, the direction of this discontinuity can easily be reversed by simply replacing $\mathds{1}_{\{g(\mathbf{x}) >  g(\mathbf{x}^*_n) \}}$ with $\mathds{1}_{\{g(\mathbf{x}) \geq  g(\mathbf{x}^*_n) \}}$ in the TVR utility \eqref{eq:util}; this would result in an axis of discontinuity that spikes \textit{up} along $\mathbf{x} = \mathbf{x}_n^*$. However, such a change should intuitively have little practical effect on the method, as $g(\cdot)$ is continuous and is not directly observed; thus, the direction of this discontinuity seems quite arbitrary. Second, from a practical perspective, there is little justification for introducing discontinuities in the acquisition function by significantly down-weighting or up-weighting the sampling of points along the axis $\mathbf{x} = \mathbf{x}_n^*$, particularly given the arbitrary nature of its direction. 

To bypass this issue, we make use of a slight extension of the TVR acquisition:
\begin{equation}
\text{TVR}'(\mathbf{x},\boldsymbol{\theta}) = \begin{cases}
\text{TVR}(\mathbf{x},\boldsymbol{\theta}), & \quad \mathbf{x} \neq \mathbf{x}_n^*,\\
\lim_{\mathbf{x} \rightarrow \mathbf{x}_n^*} \text{TVR}(\mathbf{x},\boldsymbol{\theta}) = 0.5 \text{VR}_n(\mathbf{x},\boldsymbol{\theta}), & \quad \mathbf{x} = \mathbf{x}_n^*,
\end{cases}
\label{eq:tvrmod}
\end{equation}
where $\text{TVR}(\mathbf{x},\boldsymbol{\theta})$ is as defined in \eqref{eq:tvr}. With this, one can easily show that $\text{TVR}'(\mathbf{x},\boldsymbol{\theta})$ is continuous and differentiable, thus allowing for effective optimization of this acquisition function via gradient-based methods \citep{nocedal1999numerical}. In our later implementation, we made use of the \texttt{BoTorch} library \citep{balandat2020botorch} in Python for maximizing
$\text{TVR}'(\mathbf{x},\boldsymbol{\theta})$, which seems to work quite well in selecting the next point $(\mathbf{x}_{n+1},\boldsymbol{\theta}_{n+1})$.

Another tool we found to be useful for TVR acquisition optimization is automatic differentiation \citep{baydin2018automatic}, a rapidly developing area in computer algebra. Automatic differentiation is a family of techniques that facilitate the automatic and accurate calculation of numerical derivatives via symbolic rules of differentiation. This automatic computation of derivatives circumvents the need to analytically derive gradients of highly complex expressions by hand, allowing for the easy use of gradient information in optimization with little additional effort by the user. Since $\text{TVR}'(\mathbf{x},\boldsymbol{\theta})$ is differentiable, we are able to make use of automatic differentiation for reliable optimization of the acquisition, with multiple restarts of the procedure for global optimization. In contrast, the acquisition function proposed by \cite{frazier-TP-2022} admits no closed-form expression for its gradient, and thus it is difficult to leverage automatic differentiation for effective acquisition optimization there. Instead, expensive Monte Carlo approximations must be employed, which may lead to a mediocre selection of subsequent points. 

\subsection{Batch TVR}
\label{sec:batch}
% \cmtS{maybe add a few thoughts on here? would be important for practical implementation (and thus technometrics). can simply be a simple extension of what people have done for batched EI. can show this in one simulation setting (no need for all).} \cmtJ{Responded. Filled this seciton in below.}
In practical applications, one can often perform multiple evaluations of $f$ simultaneously, particularly with the advent of parallel and distributed computing systems. We thus present a batched extension of the TVR, for querying a batch of $k \geq 2$ evaluation points from the simulator. Let us define the following $k$-TVR acquisition function:
\begin{equation}
\text{$k$-TVR}(\mathcal{D}_{(k)}) = \mathbb{E}\left[\mathds{1}_{\left\{\underset{i = 1, \cdots, k}{\max} g(\mathbf{x}_{(i)})>g(\mathbf{x}_n^*) \right\}}\text{VR}_{n,k}\left(\argmax_{i = 1, \cdots, k} g(\mathbf{x}_{(i)});\mathcal{D}_{(k)}\right)\right].
\label{eq:ktvr}
\end{equation}
Here, $\mathcal{D}_{(k)}=\{(\mathbf{x}_{(1)},\boldsymbol{\theta}_{(1)}), \cdots,(\mathbf{x}_{(k)},\boldsymbol{\theta}_{(k)}) \}$ is the set of $k$ points to potentially evaluate next from the simulator. The function $\text{VR}_{n,k}(\mathbf{x};\mathcal{D}_{(k)})$ is a natural extension of the aforementioned variance reduction criterion \citep{gramacy2015local} for the batched setting:
\begin{equation}
    \text{VR}_{n,k}(\mathbf{x};\mathcal{D}_{(k)})=\text{Var}[g(\mathbf{x})|\mathbf{f}_n]-\text{Var}\left[g(\mathbf{x})| \mathbf{f}_n, \left\{f(\mathbf{x}',\boldsymbol{\theta}')\right\}_{(\mathbf{x}',\boldsymbol{\theta}')\in\mathcal{D}_{(k)}}\right].
    \label{eq:kvr}
\end{equation}
This quantifies the reduction in variance for objective $g(\mathbf{x})$ provided additional evaluations of $f$ at $\mathcal{D}_{(k)}$. As before, the two variance terms in \eqref{eq:kvr} admit closed-form expressions from \eqref{eq:postg2} and \eqref{eq:postgcl2}. The intuition for this batched TVR acquisition is also directly analogous to the earlier (non-batched) TVR: we wish to sample points that jointly lie within the region of improvement and offer large variance reduction on the desired objective $g$.

The proposition below provides a closed-form expression for this batch TVR acquisition:
\begin{proposition}
Suppose there are no duplicates in the set of points $\{\mathbf{x}_{(1)}, \cdots, \mathbf{x}_{(k)}, \mathbf{x}_n^*\}$. Given $k \geq 2$, the \textup{$k$-TVR} acquisition function in \eqref{eq:ktvr} can then be simplified as:
\begin{equation}
\textup{$k$-TVR}(\mathcal{D}_{(k)}) = \sum_{i=1}^k p_{i,n} \textup{VR}_{n,k}\left(\mathbf{x}_{(i)};\mathcal{D}_{(k)}\right), \quad p_{i,n} =  \prod_{j=1}^{k-1} \left(1 - \Phi\left(-\alpha_{i,j}\right)\right),
\label{eq:kvr2}
\end{equation}
where $\Phi$ is the standard normal c.d.f., and:
\begin{equation}
\boldsymbol{\alpha}_i = (\alpha_{i,j})_{j=1}^k = \left(\mathbf{A}_i \boldsymbol{\Sigma}_{n,k} \mathbf{A}_i^\top\right)^{-\frac{1}{2}} \mathbf{A}_i\boldsymbol{\mu}_{n,k}.
\end{equation}
Here, $\boldsymbol{\mu}_{n,k} = (\mu_n(\mathbf{x}_{(1)}),\cdots,\mu_n(\mathbf{x}_{(k)}),\mu_n(\mathbf{x}_n^*))^\top$ is the posterior mean of $g(\cdot)$ at the new points $\mathbf{x}_{(1)}, \cdots, \mathbf{x}_{(k)}$ and the current solution $\mathbf{x}_n^*$,  $\boldsymbol{\Sigma}_{n,k} = [s_n(\mathbf{x},\mathbf{x}')]_{\mathbf{x},\mathbf{x}'}$ is the posterior covariance matrix for the same $k+1$ points, and $\mathbf{A}_i \in \mathbb{R}^{k\times (k+1)}$ is a sparse matrix with $(u,v)$-th entry:
\begin{equation}
\mathbf{A}_{i,uv} = \begin{cases}
1, & v = i,\\
-1, & u = v < i \textup{\quad or \quad } v = u + 1 > i,\\
0 & \textup{otherwise}.
\end{cases}
\end{equation}

\label{prop:prob_calc}
\end{proposition}
% $A(i)_{ji} = 1$ for $j=1,2,\cdots,k$ and $A(i)_{p,p} = -1$ for $1\leq p < i$, and $A(i)_{p-1,p}=-1$ for $i<p\leq k$.
% \cmtS{can we write the eqn for $p$ directly for the problem? i commented out previous writing below - feel free to use. we can defer proof in supplementary materials.} \cmtJ{Added. Proof in appendix}
\noindent This closed-form acquisition again permits the use of automatic differentiation and gradient-based methods for batch TVR optimization. It also performs a \textit{joint} selection of control and noise parameters for the $k$ potential points, thus allowing for effective use of fitted control-to-noise interactions for robust optimization.

For Proposition \ref{prop:prob_calc}, the requirement of no duplicates in $\{\mathbf{x}_{(1)}, \cdots, \mathbf{x}_{(k)}, \mathbf{x}_n^*\}$ is needed to maintain the positive-definiteness of $\boldsymbol{\Sigma}_{n,k}$ for computing the probabilities $\{p_{i,n}\}$. This is very much akin to the earlier discontinuity issue for the original TVR (see Section \ref{sec:disc}). We can thus use a similar strategy of defining a slightly modified batch TVR acquisition $k\text{-TVR}'(\mathbf{x},\boldsymbol{\theta})$, by extending the acquisition function \eqref{eq:ktvr} via its limit over its axes of discontinuities (i.e., in the presence of duplicates). Further details on this slight extension is provided in Supplementary Materials.

\subsection{Algorithm statement}
\label{sec:alg}
% \cmtS{write out sampling procedure as an algorithm (initial design, fitting GP, TVR optimization, etc). add a few paragraphs describing some of these steps (e.g., initial design, acquisition optimization, model fitting, etc) in more detail.} \cmtJ{Responded. Filled out this section below.}

\begin{algorithm}[!t]
    \caption{Targeted Variance Reduction for robust Bayesian optimization}
     \textbf{Input}: Expensive simulator $f(\mathbf{x},\boldsymbol{\theta})$, maximum evaluations $N$.
    \begin{algorithmic}[1]

        \State $\bullet$ Evaluate simulator at initial design points $\{(\mathbf{x}_i,\boldsymbol{\theta}_i)\}_{i=1}^{n_{\text{init}}}$, yielding outputs $\mathbf{f}_{n_{\text{init}}}$.
        
        \For{$n = n_{\text{init}},\cdots,N-1$}
            \State $\bullet$ Fit the GP model in Section \ref{sec:tvrgp} using data $\mathbf{f}_n$.
            \State $\bullet$ Optimize next evaluation point $(\mathbf{x}_{n+1},\boldsymbol{\theta}_{n+1}) \leftarrow \argmax_{(\mathbf{x},\boldsymbol{\theta})} \text{TVR}_n'(\mathbf{x}, \boldsymbol{\theta})$.
            \State $\bullet$  Evaluate the simulator $f$ at next point $(\mathbf{x}_{n+1},\boldsymbol{\theta}_{n+1})$.
            \State $\bullet$ Update data $\mathbf{f}_{n+1} = [\mathbf{f}_n; f(\mathbf{x}_{n+1},\boldsymbol{\theta}_{n+1})]$.
        \EndFor
        
   \noindent \hspace{-0.8cm} \textbf{Output}: Predicted solution $\mathbf{x}_N^* = \argmax_{\mathbf{x}\in\mathcal{X}}\mu_N(\mathbf{x})$.
    \end{algorithmic}
    \label{algorithm:TVR}
\end{algorithm}

To summarize, we provide a full algorithm statement of the TVR, with details on initial design, model fitting and acquisition optimization. Algorithm \ref{algorithm:TVR} summarizes these steps.

% The procedure for solving \eqref{eq:form} using Bayesian optimization with the TVR acquisition function is given by Algorithm \ref{algorithm:TVR}. We discuss the details of this algorithm below.

% \subsubsection{Initial Design}
Consider first the set-up of initial design points $\{(\mathbf{x}_i,\boldsymbol{\theta}_i)\}_{i=1}^{n_{\rm init}}$. While our method can be used even if such data were not designed, a careful design of initial points facilitates better initial learning of the response surface $f$ and thus of the objective $g$. One desirable property of initial design points is it should be well-spread over the joint space of control and noise parameters; there has been much work on this in the context of Latin hypercube designs \citep{mckay2000comparison}. However, we also need to accommodate for the noise distribution on $\boldsymbol{\Theta}$, which may be highly non-uniform. To account for this, we make use of the following initial designs in later numerical experiments. We first generate a random $n_{\rm init}$-point Latin hypercube design over the unit hypercube $[0,1]^{d+q}$ (where $d$ and $q$ are the number of control and noise parameters, respectively), then perform the inverse transform $P^{-1}$ on the last $q$ parameters. This appears to yield good empirical performance in simulations.

% required to begin the Bayesian optimization procedure. In general there are no constraints on how these data are obtained. However, the choice of initial experimental design (i.e., the selection of the $\mathbf{x}_i$ and $\boldsymbol{\theta}_i$ in $\mathcal{D}_{n_\text{init}}$) can have a substantial effect on the performance of the optimization algorithm. For continuous variables, we recommend an initial experimental design exhibiting \textit{space-filling} behavior (i.e., that design points are spread across the full range of all input variables) such as Latin hypercube sampling \citep{mckay2000comparison} or a maximum projection design \citep{Joseph_maxpro}. The $\mathbf{x}_i$ and $\boldsymbol{\theta}$ can be selected jointly by such a design when $\mathcal{P}$ is continuous by sampling on the $(p+q)$-dimensional unit hypercube and later applying appropriate transformations.

% When $\mathcal{P}$ is discrete we favor utilizing independent designs for $\mathbf{x}$ and $\boldsymbol{\theta}$, which does not seem to present issues in our numerical experiments within this setting for which $p=q=1$.

% For problems with higher dimension, it may be advisable to utilize experimental design techniques specialized for mixed-variable settings.

% \subsubsection{Model Fitting}
Next, with data collected from this initial design, we then fit the presented GP model in Section \ref{sec:tvrgp}. In later experiments, we make use of the \texttt{GPyTorch} package \citep{gardner2018gpytorch} in Python for GP model training. Kernel variance and length-scale parameters are fitted via maximum a posteriori estimation \citep{pml1Book} using default priors specified in \texttt{GPyTorch}, namely, $\text{Gamma}(2,0.15)$ priors for variance parameters and $\text{Gamma}(3,6)$ priors for length-scales. This GP fitting procedure is repeated after collecting a new evaluation point (or batch of evaluation points).

% While maximum likelihood estimation or a fully Bayesian analysis would work to estimate Gaussian process kernel parameters, we opt for MAP estimation to learn these parameters.

% When $\boldsymbol{\Theta}$ is not isotropic standard Gaussian, $\tilde{f}(\mathbf{x},\mathcal{T}(\mathbf{z}))$ should be used in place of $f(\mathbf{x},\boldsymbol{\theta})$ for modeling, where $\mathcal{T}^{-1}(\boldsymbol{\Theta})\sim \mathcal{N}(\mathbf{0,I})$. 

% \subsubsection{Optimization of the Acquisition Function}
With the fitted GP in hand, the proposed TVR acquisition \eqref{eq:tvrmod} is implemented using the \texttt{PyTorch} package \citep{paszke2019pytorch} in Python. To maximize this acquisition, we make use of the \texttt{BoTorch} library \citep{balandat2020botorch} in Python, which employs automatic differentiation \citep{baydin2018automatic} for gradient calculation, and standard nonlinear optimization methods in \texttt{SciPy} \citep{2020SciPy-NMeth}. Since the TVR acquisition is not convex, we made use of multiple restarts of the optimization procedure (with different initial starting points) to better tackle the global maximization problem. We found that five restarts appear to strike a good balance between optimization performance and computational efficiency.

Finally, with the new point $(\mathbf{x}_{n+1},\boldsymbol{\theta}_{n+1})$ optimized, we then evaluate the simulator $f(\cdot)$ at this new point, and refit the underlying GP model. This procedure is then repeated until either a budget is reached on the number of simulator evaluations, or an acceptable objective is achieved on $g$. Recall that the objective $g$ is not directly observed from function evaluations on $f$. In later experiments, we use the predicted optimal solution $\mathbf{x}_n^* = \argmax_{\mathbf{x} \in \mathcal{X}} \mu_n(\mathbf{x})$, where $\mu_n(\mathbf{x}) = \mathbb{E}[g(\mathbf{x})|\mathbf{f}_n]$ is the surrogate point prediction of the objective $g(\mathbf{x})$.

% We perform sequential acquisitions with a fixed evaluation budget of $T-n_{\text{init}}$ evaluations of $f$. After this budget has been exhausted, the maximizer of $g$'s posterior mean (conditioned on $\mathcal{D}_T$) is returned as the terminal solution. However, any risk measure of $g$'s distribution could be used in place of posterior mean depending on the user's risk tolerance.

\section{Numerical Experiments}
\label{sec:num}

% The first is a custom trigonometric numerical test function, treating one input as a control variables and the other input as a noise parameter. The second is the Trid 6D numerical test function, treating three dimensions as control variables and the remaining three dimensions as noise parameters.

We now investigate the effectiveness of TVR compared to the state-of-the-art in a suite of simulation experiments with varying difficulties. All methods begin with the same initial design points $\{(\mathbf{x}_i,\boldsymbol{\theta}_i)_{i=1}^n\}$, obtained via the procedure in Section \ref{sec:alg}. The compared methods are described below:
\begin{itemize}[leftmargin=*]
    \item \textit{Random design}: Subsequent design points $(\mathbf{x}_{n+1},\boldsymbol{\theta}_{n+1}), (\mathbf{x}_{n+2},\boldsymbol{\theta}_{n+2}), \cdots$ are independently sampled, uniformly-at-random for control parameters $\mathbf{x}$ and from the noise distribution $\mathcal{P}$ for the noise parameters $\boldsymbol{\theta}$.
    \item \textit{Two-stage design}: Subsequent design points are selected via the two-stage procedure in \cite{Williams2000,Groot2010BayesianMC}. First, $\mathbf{x}_{n+1}$ is selected to maximize EI on the objective $g$ (see \eqref{eq:stg1}); a slight modification was used for efficiency (see Supplementary Materials). Next, $\boldsymbol{\theta}_{n+1}$ is selected to minimize mean-squared prediction error at the chosen $\mathbf{x}_{n+1}$ (see \eqref{eq:stg2}). The GP model is then refit, and this procedure is repeated for subsequent design points. 
    \item \textit{Variance reduction}: Subsequent points are selected via the variance reduction criterion \citep{gramacy2015local} on objective $g$, i.e.:
    \begin{equation}
(\mathbf{x}_{n+1},\boldsymbol{\theta}_{n+1}) \leftarrow \argmax_{(\mathbf{x},\boldsymbol{\theta})}\text{VR}_n(\mathbf{x},\boldsymbol{\theta}) = \text{Var}[g(\mathbf{x})|\mathbf{f}_n]-\text{Var}[g(\mathbf{x})| \mathbf{f}_n, f(\mathbf{x},\boldsymbol{\theta})],
    \label{eq:v_acq}
    \end{equation}
where as before, both variance terms admit closed-form expressions from \eqref{eq:postg2} and \eqref{eq:postgcl2} from our model in Section \ref{sec:tvrgp}. While this is not an existing method specifically developed for robust optimization, we include this benchmark to gauge how effective our \textit{targeting} of variance reduction is within the region of improvement for the TVR.
    \item \textit{Knowledge gradient}: Subsequent points are selected via the knowledge gradient approach in \cite{frazier-TP-2022} for the robust optimization problem \eqref{eq:form}. Here, we used the default number of 64 Monte Carlo samples to approximate the acquisition function, as recommended for the standard knowledge gradient in the \texttt{BoTorch} package.
    % Subsequent $\mathbf{x}$ points are selected by maximizing the knowledge gradient \citep{knowledgegradient2008} acquisition function, while the associated $\boldsymbol{\theta}$ values are randomly sampled from their distribution $\mathcal{P}$. This is equivalent to treating $\boldsymbol{\theta}$'s impact on $g$ as Gaussian noise when making acquisitions. Here, the terminal solutions $\mathbf{x}^*_n$ are still returned as the maximizers of $\mu_n$ as given by \eqref{eq:postgcl2}. The implementation of the extension of the knowledge gradient to the robust optimization problem \eqref{eq:form} provided by \cite{frazier-TP-2022} cannot accommodate our preferred initial experimental design. As a result, we exclude it from our experiments to maintain a fair comparison between the evaluated methods.
    % \cmtJ{For now, what I have as knowledge gradient is to simply use the standard knowledge gradient with respect to x and randomly sample $\theta$. This is to demonstrate performance in the case where you simply treat the influence of theta as Gaussian noise when making acquisitions (but are allowed to use $\theta$'s distribution when returning a final answer). I'll try to get Frazier's student's code working to include their robust opt method (which is not what we have currently). I'll write the description for what the current results are for below:} 
\end{itemize}
For all methods, the predicted solution $\mathbf{x}_n^*$ is selected as the solution $\mathbf{x}$ that maximizes the posterior mean $\mu_n(\mathbf{x})$ in \eqref{eq:postgcl2}. All methods are replicated for 100 trials to show simulation variability with respect to initial design points. All methods also share the same randomly-generated initial design within each trial. For fair comparison, each method is provided a fixed amount of computation time for evaluation of its acquisition function.

% (3 seconds in Section \ref{sec:trig}, 2 seconds in Section \ref{sec:trid}, and 1.5 seconds in Section \ref{sec:app}).

% We compare TVR's performance against that of four acquisition strategies: (1) variance reduction, (2) random sampling, (3) knowledge gradient (use KG to choose $x_{new}$ and then sample $\theta_{new}$ from its distribution), (4) sequential selection (use expected improvement on $G$ to choose $x_{new}$ and then use variance reduction to select $\theta_{new}$). As opposed to the knowledge gradient approach used in the example depicted in Figure \ref{fig:trig_TVR_demo}, we consider the problem structure (i.e., that the objective $G$ is the expected value of observable function $f$) when choosing a ``terminal" solution (after each function evaluation) with all approaches. For each method, the point $x$ maximizing the posterior mean $\mu_t(\cdot)$ is returned. Within each trial of each problem, all five acquisition functions share the same initial set of observations to ensure a fair comparison. Each experiment is repeated for 100 trials. Figures \ref{fig:trigonometric_test}, \ref{fig:Trid}, and \ref{fig:braking} show mean performance for each method, along with 10-th and 90-th quantiles.

\subsection{1D-1D trigonometric function}
\label{sec:trig}

\begin{table}[!t]
\centering
\begin{tabular}{ c c c c c c c}
\toprule 
\multicolumn{7}{c}{Noise distribution 1}\\
 \hline
 $m$&-1&-2/3&-1/3&1/3&2/3&1\\
$p_m = \mathbb{P}(\Theta=m)$&.2088&.1612&.0792&.0811&.1137&.3561\\
 \toprule
\end{tabular}
\vspace{0.25cm}
\begin{tabular}{ c c c c c c c}
 \toprule
\multicolumn{7}{c}{Noise distribution 2}\\
\hline
$m$&1/2&8/15&17/30&3/5&19/30&2/3\\
$p_m = \mathbb{P}(\Theta=m)$&.0762&.2509&.1454&.2080&.1057&.2138\\
 \toprule
\end{tabular}
\caption{Probability mass functions for the two considered noise distributions in the 1D-1D trignometric function experiment.}
\label{tbl:dist}
\end{table}

\begin{figure}
\centering
\begin{subfigure}{.5\textwidth}
  \centering
  \includegraphics[width=1.0\linewidth]{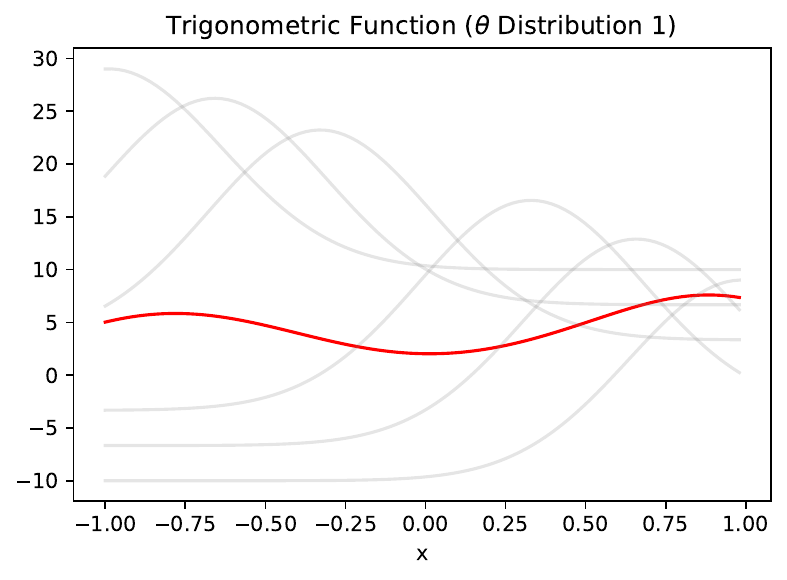}
\end{subfigure}%
\begin{subfigure}{.5\textwidth}
  \centering
  \includegraphics[width=1.0\linewidth]{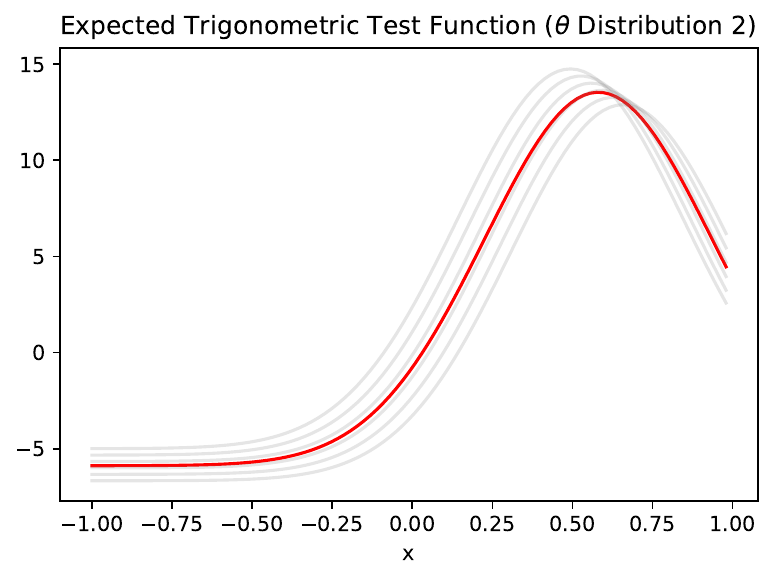}
\end{subfigure}
\caption{Visualizing the trignometric test function $f(x,\theta)$ for different sample draws of $\Theta$ from noise distribution 1 (left) and 2 (right). Plotted in red is the desired objective function $g(x) = \mathbb{E}[f(x,\Theta)]$ to maximize.}
\label{fig:trigshape}
\end{figure}

We consider first the following two-dimensional trigonometric test function, with $d=1$ control and $q=1$ noise parameter:
\begin{equation}
    f(x,\theta) = 2\cos\left(\frac{x}{\pi} \right)\exp\left\{-4 (x-\theta)^2 \right\}-\theta.
\end{equation}
Here, the design space for the control parameter is $x\in[-1,1]$. We explore two simple discrete choices of noise distributions for $\boldsymbol{\Theta}$ on $[0,1]$, summarized in Table \ref{tbl:dist}. The first distribution is chosen such that there is large probability mass on regions with control-to-noise interactions. The second is chosen to emphasize regions with less control-to-noise interactions. Figure \ref{fig:trigshape} visualizes this difference by showing sample paths of $f(x,\Theta)$ for different draws of $\Theta$. For the first distribution, we see that the presence of interactions results in large variation in the maximizer over $\mathbf{x}$; for the second distribution, the lack of interactions results in little variation of the maximizer. All methods begin with $n=10$ initial design points, then proceed with 20 sequential design points.

% For each experiment an appropriate grid of values is chosen, and the probability mass function is sampled from a Dirichlet distribution. For the first, a distribution is chosen such that there is moderate variation in $f$ due to $\theta$:

% For the second, a distribution is chosen such that there is limited variability in $f$ due to $\theta$ and the location of the minimum changes only slightly across possible values of $\theta$:

\begin{figure}[!t]
\centering
\begin{subfigure}{.5\textwidth}
  \centering
  \includegraphics[width=1.0\linewidth]{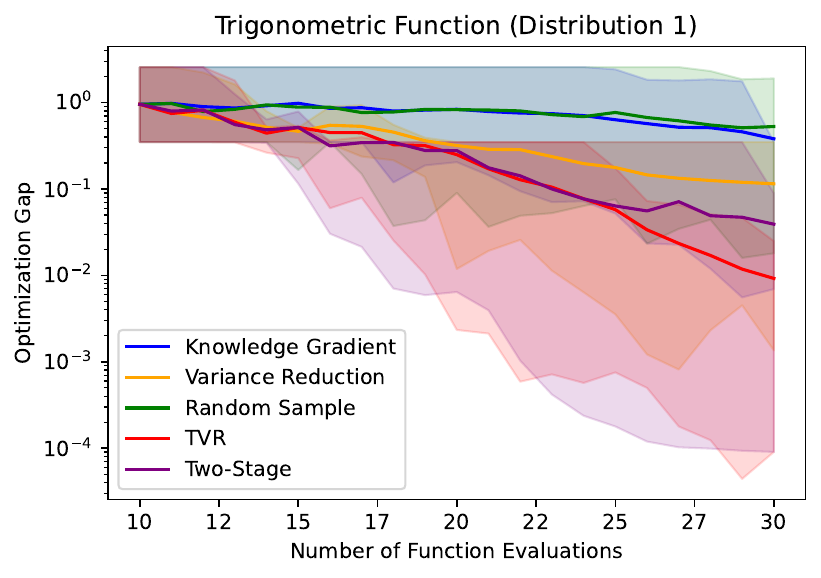}
\end{subfigure}%
\begin{subfigure}{.5\textwidth}
  \centering
\includegraphics[width=1.0\linewidth]{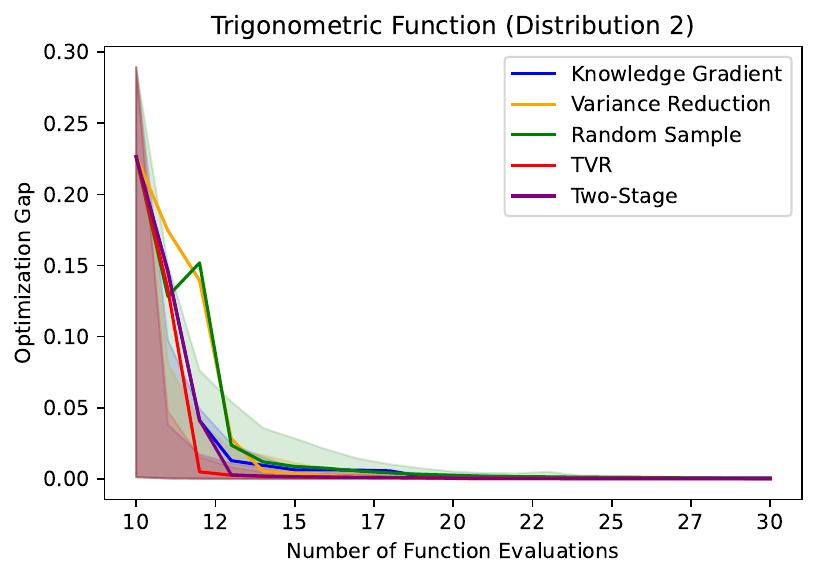}
\end{subfigure}
\caption{Plotting the optimization gap $g(\mathbf{x}^*) - g(\mathbf{x}_n^*)$ against the number of function evaluations on $f$ for the 1D-1D trignometric function experiment. The solid lines mark the average optimization gap over 100 trials, and the shaded regions mark its 10th and 90th quantiles.}
\label{fig:trigonometric_test}
\end{figure}

Figure \ref{fig:trigonometric_test} shows the optimization performance of the compared methods for both choices of noise distribution. We see that the TVR yields better performance (in terms of a smaller optimization gap) than the two-stage approach for both distributions. 
This is not too surprising, as the TVR can better exploit the underlying control-to-noise interactions via the use of a \textit{joint} acquisition function for choosing next query points. This improvement can be seen even when such interactions are more subtle, i.e., for distribution 2. 
The TVR also improves upon the remaining three methods (random sampling, variance reduction and knowledge gradient). This suggests several things. First, compared to variance reduction, the \textit{targeting} of variance reduction within the improvement region seems to be effective via the new exploration-exploitation-precision trade-off. Second, compared to the knowledge gradient approach, the use of a closed-form acquisition (with automatic differentiation) appears to be a contributing factor for higher quality sequential points for robust optimization.

% Figure \ref{fig:trigonometric_test} summarizes optimization performance of the various methods for both $\theta$ distributions. TVR and Sequential Acquisition perform comparably in both experiments. It seems that when problem dimension is low and the noise is relatively simple, it is not necessary to jointly select $x$ and $\theta$ when choosing design points. In this simple case, choosing $x$ using a standard acquisition function and then choosing $\theta$ suffices. 

\subsection{3D-3D Trid function}
\label{sec:trid}

\begin{figure}[!t]
    \centering
    \includegraphics[width=.4\textwidth]{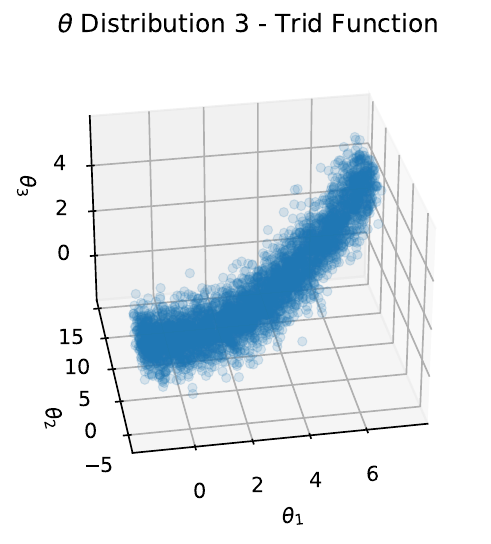}
    \caption{Visualizing samples drawn from the complex correlated distribution $\Theta$ in the 3D-3D Trid function experiment.}
    \label{fig:trigNFnoise}
\end{figure}

\begin{figure}[!t]
\centering
\begin{subfigure}{.48\textwidth}
  \centering
  \includegraphics[width=1.0\linewidth]{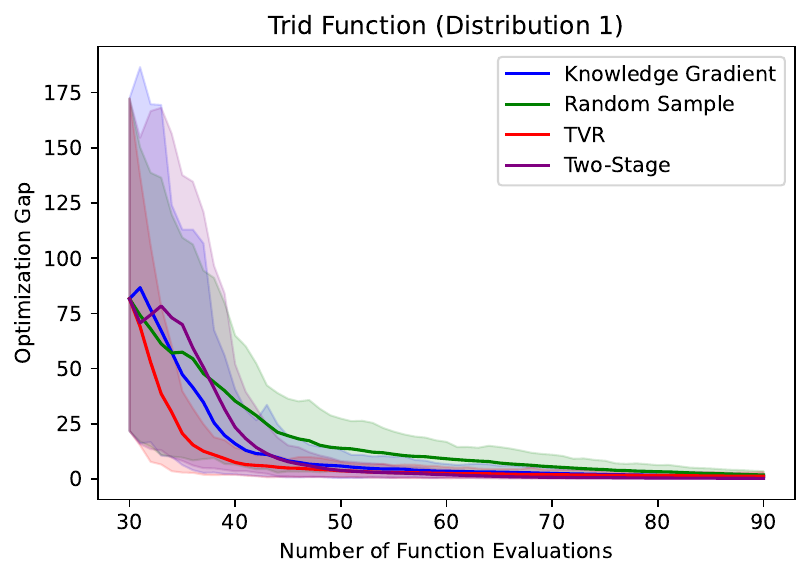}
\end{subfigure}%
\begin{subfigure}{.48\textwidth}
  \centering
  \includegraphics[width=1.0\linewidth]{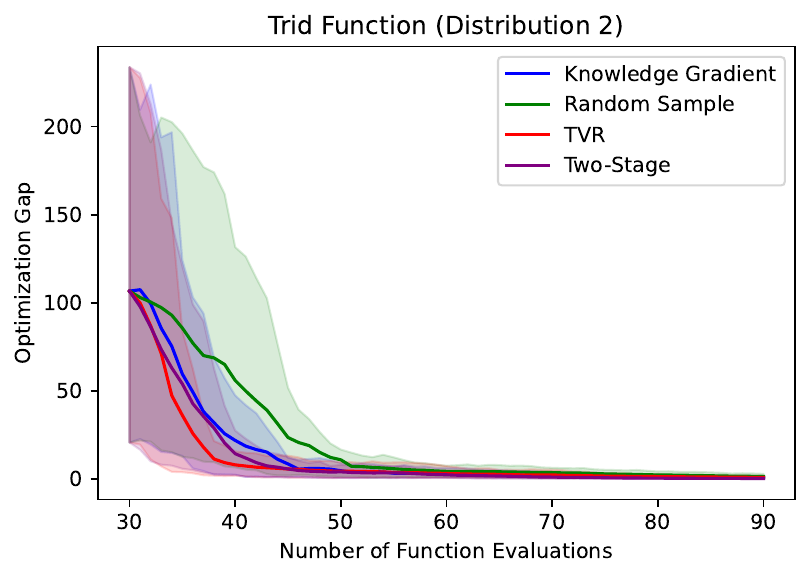}
\end{subfigure}
\begin{subfigure}{.48\textwidth}
  \centering
  \includegraphics[width=1.0\linewidth]{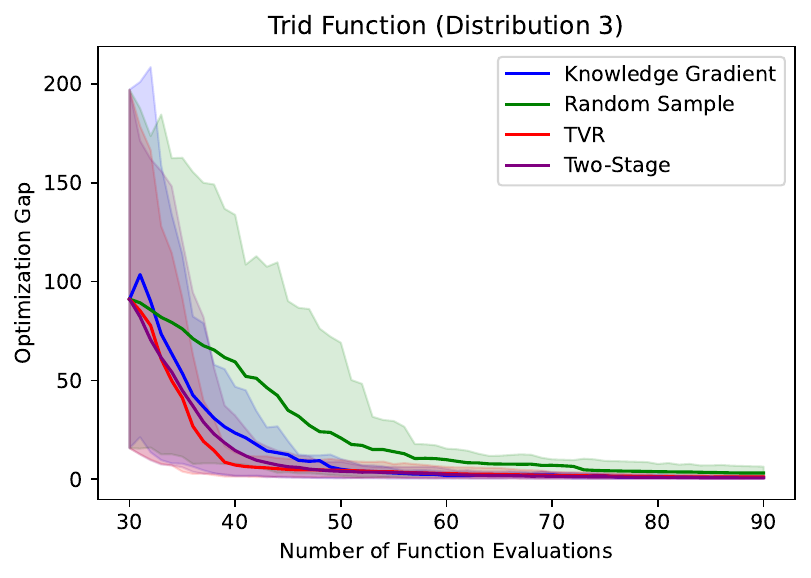}
\end{subfigure}

% \caption{Optimization for the Trid 6D test function, for each of the three chosen distributions on $\theta$. Results from Variance Reduction are omitted from the first two sets of results due to exceptionally poor performance. All methods achieve comparable performance when the evaluation budged is exhausted or near-exhausted, but TVR performs best on average when the number of function evaluations is small for all three distributions on $\theta$.}
\caption{Plotting the optimization gap $g(\mathbf{x}^*)-g(\mathbf{x}_n^*)$ against the number of function evaluations on $f$ for the 3D-3D Trid function experiment. The solid lines mark the average optimization gap over 100 trials, and the shaded regions mark its 10th and 90th quantiles. The variance reduction method is omitted in these plots due to its large optimization gap.}
\label{fig:Trid}
\end{figure}

We then consider the higher-dimensional Trid test function, taken from \cite{surjanovic2013virtual}, with $d=3$ control and $q=3$ noise parameters:
\begin{equation}
f(\mathbf{x},\boldsymbol{\theta}) = -\sum_{j=1}^6(\tau_j-1)^2 - \sum_{j=2}^6 \tau_{j}\tau_{j-1}, \quad \tau_{j} = \begin{cases}
    x_{\lfloor j/2 \rfloor +1},&j\text{ odd,}\\
    \theta_{j/2},&j\text{ even.}
\end{cases}
\label{eq:trid}
\end{equation}
Here, the design space for control parameters is $\mathcal{X
}=[-36,36]^3$. We explore the following three more complex choices of noise distributions for $\boldsymbol{\Theta}$:
\begin{itemize}[leftmargin=*]
    \item \textit{Rescaled Beta}: $\Theta_j \distas{indep.} 72\times \text{Beta}(3j,10-3j)-36, \; j = 1, \cdots, 3$,
    \item \textit{Mixed}: $\Theta_1 \sim 72\times \text{Beta}(3,7)-36$, $\Theta_2 \sim \mathcal{N}(2,4)$, $\Theta_3 \sim \text{Exp}(1/6)$, with all random variables independent,
    \item \textit{Correlated}: $\Theta_1 \sim \text{Unif}[-1.5,7.5]$, $\Theta_2|(\Theta_1 = \theta_1) \sim \mathcal{N}(\theta_1+1,4)$, $\Theta_3|(\Theta_1 = \theta_1) \sim \mathcal{N}\left( (\frac{\theta_1-1}{3})^2,\frac{1}{4}\right)$.
\end{itemize}
For the first two distributions, it is straightforward to derive via its inverse c.d.f. the required transformation $\mathcal{T}$ for fitting the GP model in Section \ref{sec:tvrgp}. The last is the most complex distribution of the three, with probability mass concentrated on a nonlinear manifold on $\mathbb{R}^3$. We thus make use of the normalizing flow approach from Section \ref{sec:tvrgp} (specifically, via the nonlinear independent components estimation approach in \citealp{dinh2015nice}) to learn the appropriate transformation $\mathcal{T}$ needed for closed-form GP modeling. All methods begin with $n=30$ initial points, then proceed with 60 sequential design points.

% In the first, $\theta_j\stackrel{D}{=} 72\times \text{Beta}(3j,10-3j)-36$. In the second, $\theta_1\stackrel{D}{=}72\text{Beta}(3,7)-36$, $\theta_2\sim N(2,4)$, and $\theta_3\sim \text{Expo}(1/6)$. In the third, $\theta$ follows a more complex unnamed distribution shown in Figure \ref{fig:trigNFnoise}. $\theta$ is generated from this distribution as follows:
% $$\theta_1 \sim\text{Unif}(-30,36)$$
% $$\theta_2 = N(\theta_1+1,9)$$
% $$\theta_3 = 10\sin\left(\frac{(\theta_1-30)\pi}{66} \right)$$

% For the thitrd distribution, a normalizing flows neural network implementing the NICE \cite{dinh2015nice} architecture is trained to learn a suitable mapping from isotropic Gaussian to the target density (inverse CDF transforms suffice in the other cases). For all three $\theta$ distributions, an appropriate mapping 
% on $\theta$ is integrated into the optimization procedure (as described in Section \ref{section:beyondGauss}) for all tested acquisition functions to enable the direct use of all derived quantities for the integral-valued GP.

Figure \ref{fig:Trid} shows the optimization performance of the compared approaches for the three noise distributions. The variance reduction approach is excluded due to its poor performance, which skews the plots when comparing the remaining methods. In this higher-dimensional experiment, the TVR now considerably outperforms the two-stage approach for all noise distributions. One reason for this is that, given the limited number of evaluations on a higher-dimensional space (both for control and noise parameters), the exploitation of control-to-noise interactions becomes more important for identifying robust control settings. The TVR better achieves this via a joint acquisition function over $\mathbf{x}$ and $\boldsymbol{\theta}$, in contrast to the two-stage approach, which performs separate selection of control and noise parameters under different acquisitions. As before, the TVR also greatly improves upon random sampling and variance reduction; this suggests the importance of targeting for variance reduction. Our approach again improves upon the knowledge gradient approach \citep{frazier-TP-2022}. Thus, when control-to-noise interactions are present in the simulated response surface $f$, the TVR appears to offer considerable improvements over existing methods.

\begin{figure}[!t]
    \centering
    \includegraphics[width=.5\textwidth]{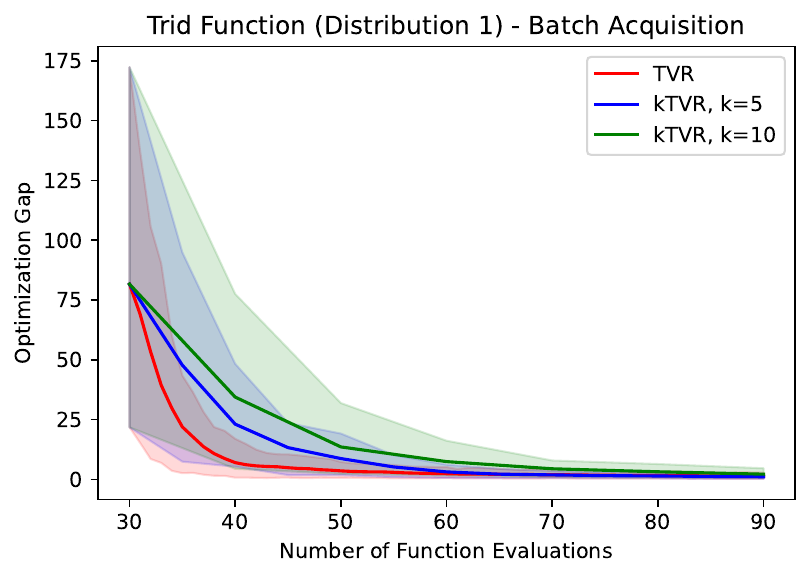}
    \caption{Plotting the optimization gap $g(\mathbf{x}^*)-g(\mathbf{x}_n^*)$ against the number of function evaluations on $f$ for the 3D-3D Trid function experiment, for the batch $k$-TVR with different batch sizes $k$. The solid lines mark the average optimization gap over 100 trials, and the shaded regions mark its 10th and 90th quantiles.}
    \label{fig:trid_batch}
\end{figure}

% cannot less information is available to select each subsequent design point
Finally, we investigate the batch $k$-TVR approach (Section \ref{sec:batch}) using noise distribution 1. Figure \ref{fig:trid_batch} shows the optimization performance of the $k$-TVR with batch sizes of $k=5$ and $k=10$, compared with the purely sequential TVR. As batch size $k$ increases, we see a slight deterioration in optimization performance. This is not surprising: compared to the fully sequential setting, the batch $k$-TVR would have less evaluation data points to work with when selecting subsequent design points. This slightly worse performance, of course, is offset by the fact that many simulation evaluations can be performed simultaneously, e.g., via parallel or distributed computing systems.

% Note, however, that the ability to perform batch acquisitions will often facilitate the ability to perform a greater total number of evaluations as compared to a fully sequential approach (e.g., if parallel resources are available and the Bayesian optimization procedure must be carried out under significant time pressure).

% (standard variance reduction is excluded for the first two experiments due to extremely poor performance). In this example there is clear interaction between $x$ and $\theta$, and it can be observed that $\theta$'s distribution will impact the location of the optimum $x^*$ (minimizing the corresponding $G$). Especially when the number of function evaluations is low, TVR outperforms the other methods on average. With higher problem dimension and more complex noise, there appears to be a benefit to joint selection of $x_{new}$ and $\theta_{new}$ through the use of TVR.

\section{Robust Design of Automobile Brake Discs}
\label{sec:app}

Finally, we explore an application of the TVR for the robust design of automobile brake discs under operational uncertainty. The prevention of brake failures is an integral part of automobile design, as these failures can cause significant and potentially fatal safety risks. A primary cause for such failures is brake overheating \citep{dewanto2018new}, which causes braking pad glazing and braking fluid heating, and subsequently a considerable lowering of friction force required for timely braking. There has thus been much work on designing brake disc material that is resistant to heating \citep{Ilie_braking,xiao_brakingfraction,maleque12010material}. One key challenge in this design problem is the \textit{robustness} of such brakes to uncontrollable conditions in operation, e.g., initial velocity or braking time of vehicle \citep{xiao_brakingfraction}. Given some prior knowledge on the nature of these uncontrollable factors, the goal is thus the robust design of brakes in the presence of such uncertainties.

% Automobile brake discs begin to function poorly at high temperatures, posing significant safety risks.

In this application, we aim to design the material properties of the brake disc, such that the maximum temperature reached during braking (i.e., spatially over the brake disc and temporally over the braking period) is minimized. The considered vehicle is an ordinary four-wheeled 2,000 kg car, for which braking is applied on eight brake pads. We investigate in particular three material properties of the brake disc for control parameters: thermal conductivity (denoted as $x_1$, where $x_1 \in [50,200]$ W/(m$\cdot$K)), mass density ($x_2$, where $x_2 \in[7000,9000]$ $\text{kg}/\text{m}^3$) and specific heat ($x_3$, where $x_3\in[300,700]$ J/(kg$\cdot$K)). For noise parameters, we consider two factors: the initial velocity of the vehicle at braking time ($\theta_1$, measured in m/s), and the braking time needed to a full stop ($\theta_2$, measured in seconds). As a proof of concept, we assign the following independent noise distributions: $\theta_1 \sim \mathcal{N}(27.8,1.0)$ and $\theta_2 \sim \mathcal{N}(2.75,0.25)$. It is known that both noise parameters $\theta_1$ and $\theta_2$ exhibit significant \textit{interactions} with the braking friction coefficient, since this coefficient depends on the choice of brake disc materials \citep{xiao_brakingfraction}; this thus provides an appealing case study for robust optimization. Letting $f(\mathbf{x},\boldsymbol{\theta})$ be the simulated maximum temperature at parameters $(\mathbf{x},\boldsymbol{\theta})$, our goal is to minimize the objective $g(\mathbf{x}) = \mathbb{E}[f(\mathbf{x},\boldsymbol{\Theta})]$, the maximum temperature at brake design $\mathbf{x}$, averaged under uncertainty on $\boldsymbol{\Theta}$. Note that we are minimizing rather than maximizing $g$; this can easily accommodated by using $-g$ in \eqref{eq:form}.

% Initial braking velocity and braking pressure (which is inversely related to braking time) can be highly impactful on brake wear and heat accumulation. 

% We seek to design brake disc materials that will minimize the maximum temperature reached by the brake disc during braking. We aim to select the thermal conductivity, mass density, and specific heat of the brake disc material which achieve this goal. Variation in the car's braking behavior will also affect the temperature reached by the disc. In particular, the initial velocity of the vehicle at braking time ($V_0$, measured in $m/s$) and the time spent braking from the initial velocity to a full stop ($t_\text{brake}$, measured in seconds). We account for this variation by treating $V_0$ and $t_\text{brake}$ as noise parameters in the optimization procedure, taking $V_0\sim N(27.8,1)$ and $t_\text{brake} \sim N(2.75,0.25)$. 

\begin{figure}[!t]
    \begin{subfigure}{.44\textwidth}
  \centering
  \includegraphics[width=1.0\linewidth]{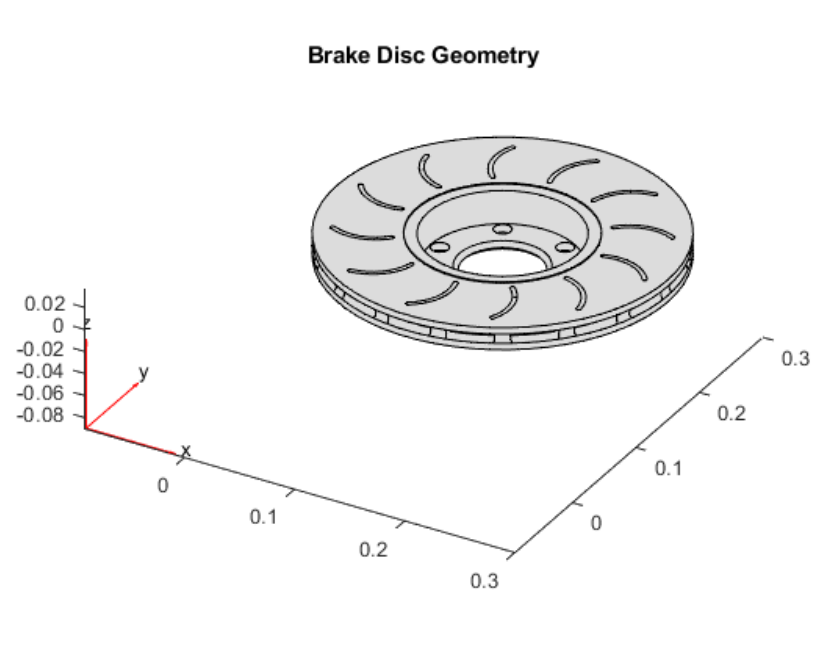}
\end{subfigure}%
\begin{subfigure}{.44\textwidth}
  \centering
  \includegraphics[width=1.0\linewidth]{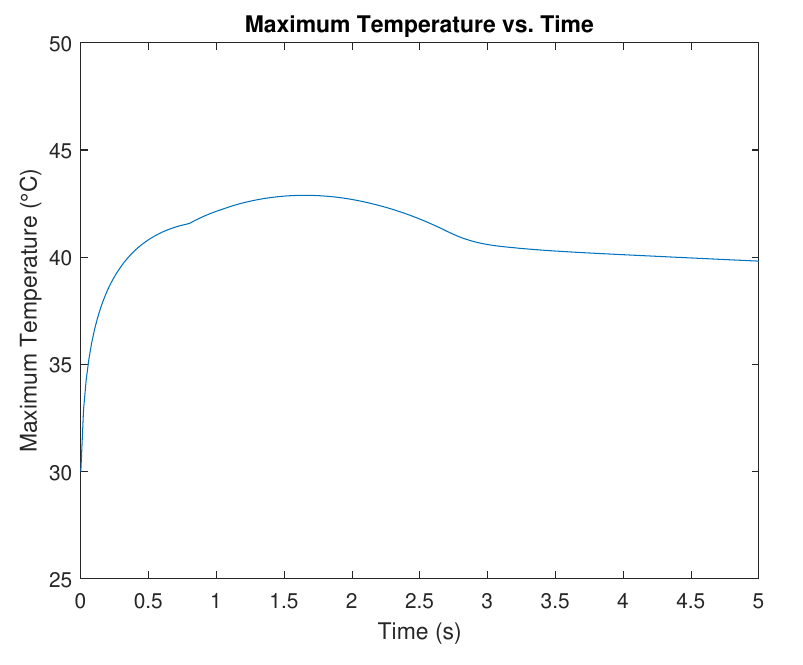}
\end{subfigure}
    
    \caption{(Left) Visualizing the 3D-modeled brake disc geometry in Matlab. (Right) Visualizing the simulated maximum temperature over the brake disc as a function of time. The desired objective to minimize, $g$, is the maximum of this profile over time.}
    \label{fig:brakevis}
\end{figure}

For such a set-up, physical experiments are clearly expensive: materials need to be designed, brakes manufactured, and braking tests performed. We thus adopt the more affordable option of computer experiments. Here, we simulate the detailed car braking procedure via the Matlab module \citep{matlab_disc}. Figure \ref{fig:brakevis} (left) shows the 3D-modeled brake disc geometry. This geometry is then used to simulate the heat generation from the brake application, then the subsequent transient heating and convective cooling in the release period. Figure \ref{fig:brakevis} (right) shows the simulated maximum temperature over the brake disc as a function of time, for a fixed choice of brake material. 

\begin{figure}[!t]
    \centering
    \includegraphics[width=.5\textwidth]{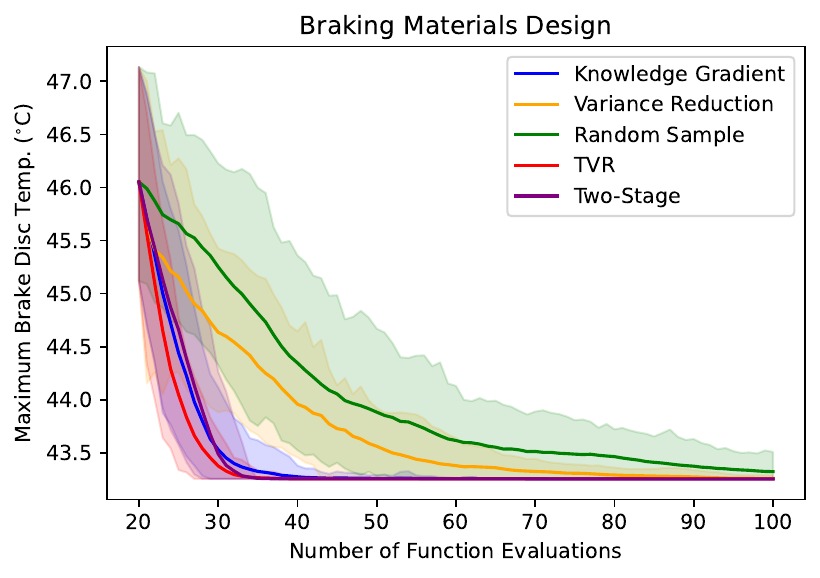}
\caption{Plotting the objective value $g(\mathbf{x}_n^*)$ of the chosen solution $\mathbf{x}_n^*$ against the number of finite element simulation runs for the automobile brake design problem. The solid lines mark the average optimization gap over 100 trials, and the shaded regions mark its 10th and 90th quantiles.}
    \label{fig:braking}
\end{figure} 

Given the fine-scale spatial and temporal resolution required in the finite element simulator, each computer simulation (at a given material design $\mathbf{x}$ and noise parameters $\boldsymbol{\theta}$) can take several minutes of computation. The desired robust optimization goal, which requires many experiments at different choices of $\mathbf{x}$ and $\boldsymbol{\theta}$, can thus be computationally costly. We compare here the TVR with the existing black-box optimization methods from before, namely the random sampling, two-stage sampling, variance reduction and knowledge gradient approaches. All methods begin with $n=20$ initial design points, then proceed with 80 sequentially selected points; this is then replicated for 100 trials. Since $g(\cdot)$ cannot be directly observed here, we would need to approximate the true objective value $g(\mathbf{x}^*_n)$ for a chosen solution $\mathbf{x}_n^*$. This is done by first running the simulator $f(\mathbf{x},\boldsymbol{\theta})$ over a separate 500-point Latin hypercube design \citep{mckay2000comparison} over $(\mathbf{x},\boldsymbol{\theta})$-space, fitting a GP surrogate on the simulated data, then approximating the desired expectation in $g(\mathbf{x}_n^*)$ via this surrogate.

Figure \ref{fig:braking} shows the objective value $g(\mathbf{x}_n^*)$ for the chosen solution $\mathbf{x}_n^*$ from each method. Results are similar to the higher-dimensional simulations in Section \ref{sec:trid}. The proposed TVR outperforms the two-stage approach, achieving lower maximum brake disc temperatures particularly at smaller sample sizes. This is not too surprising given our prior knowledge that considerable control-to-noise interactions are present in $f$; the TVR better accounts for such interactions via a joint acquisition over $\mathbf{x}$ and $\boldsymbol{\theta}$. The TVR again outperforms the remaining three approaches, which demonstrates the importance of targeted variance reduction and a closed-form acquisition function.

% To enable exact calculation of $G(\cdot)$ and accelerate computation, a Gaussian Process emulator trained on a set of simulator observations generated from a Latin Hypercube Design ($n=500$ in 5 dimensions) is used in place of the more costly simulator throughout the comparison. Figure \ref{fig:braking} displays optimization results over 100 trials. TVR outperforms the baseline methods, finding the global minimum with the fewest evaluations of $f$ on average. 

\section{Conclusion}
\label{sec:conc}

We presented a new Targeted Variance Reduction (TVR) method for robust Bayesian optimization of black-box simulators, where certain parameters are uncertain in practice and can be modeled as a ``noisy'' random variable $\Theta \sim \mathcal{P}$. The key novelty of TVR is a new acquisition function that jointly optimizes over both control parameters $\mathbf{x}$ and noise parameters $\boldsymbol{\theta}$, via the targeting of variance reduction on the objective within the desired region of improvement. Compared to existing two-stage approaches, this joint acquisition can better leverage control-to-noise interactions for effective robust optimization. We further presented a Gaussian process modeling framework on the simulated response surface, such that for a broad class of non-Gaussian noise distributions $\mathcal{P}$, the TVR admits a closed-form expression for its acquisition function. This then facilitates effective acquisition optimization for selecting the next query point. The TVR also revealed a novel exploration-exploitation-precision trade-off, which extends upon the well-known exploration-exploitation trade-off in reinforcement learning. We demonstrated the improvement of the proposed TVR over the state-of-the-art, in a suite of numerical experiments and an application to the robust design of automobile brake discs under operational uncertainty.

Given the promising results presented, there are many avenues for future work. First, it has been observed that existing Bayesian optimization approaches (and by extension, the TVR as well) may suffer from a ``curse-of-dimensionality'', in that performance may deteriorate when the parameter space becomes high-dimensional. One solution is to carefully specify a GP surrogate that can model for embedded low-dimensional structure in this high-dimensional space, see, e.g., \cite{gramacy2012Gaussian, li2023additive}, then leverage this for constructing the required acquisition function; we aim to explore this as a future direction. Another interesting direction is the extension of the TVR for broader applications beyond the physical sciences. In particular, the use of agent-based simulation models have become increasingly prevalent for guiding policy decisions \citep{ghaffarian2021agent}. Such simulators are typically expensive to perform and have many uncontrollable noise parameters, and we are currently exploring the use of the TVR for such applications.

% We have presented Targeted Variance Reduction (TVR), a novel acquisition function for Bayesian Optimization. This acquisition function allows for the efficient optimization of the expected value of functions taking two sets of arguments: design variables which can be controlled by the user ($x$) and noise parameters which are uncontrollable in practice and best treated as random ($\theta$). It readily admits a closed form and clear interpretation while dynamically balancing exploration, exploitation, and solution precision in the selection of new test points. Notably, it does so while allowing for the joint selection of $x_{new}$ and $\theta_{new}$. Through the use of variable transformations, this framework can accommodate a broad range of probability distributions over $\theta$. Furthermore, the proposed method achieves strong optimization performance on a series of test problems.

\if0\blind{
% \noindent \textbf{Acknowledgements}: The authors gratefully acknowledge funding from NSF CSSI Frameworks grant 2004571 (KL, SM), NSF DMS 2210729 (KL, SM) and U.S. Department of Energy Grant no. DE-FG02-05ER41367 (SAB, JFP).
}
\fi

% \FloatBarrier
\spacingset{1.0}
\bibliography{TVR_references}

\end{document}

\def\spacingset#1{\renewcommand{\baselinestretch}%
{#1}\small\normalsize} \spacingset{1}

\if1\blind
{
  \title{\bf Supplementary Material for ``Targeted Variance Reduction: Robust Bayesian Optimization of Black-Box Simulators with Noise Parameters"}
  % \small
  %  \author{John J. Miller\footnote{Department of Statistical Science, Duke University}\;, Simon Mak$^*$
  %  }
  \author{}
  \maketitle
} \fi

\if0\blind
{
  \bigskip
  \bigskip
  \bigskip
  \begin{center}
    {\LARGE\bf Supplementary Material for ``Targeted Variance Reduction: Robust Bayesian Optimization of Black-Box Simulators with Noise Parameters"}
\end{center}

  \medskip
} \fi

\spacingset{1.55} % DON'T change the spacing!

% \FloatBarrier

\maketitle

% \subsection{Definition of Objective Function in Section \ref{sec:motivation}}\label{sec:mot_obj}

% The objective used in the example discussed in Section \ref{sec:motivation} is defined as follows:

% \begin{align*}
% f(x,\theta) & = \left(\frac{\theta}{5}-\frac{4}{{\left( \frac{\theta^4}{2} + 1 \right)}}\right) \exp \left( -8 \left( x + \frac{\theta}{20} - \frac{8}{5} \right)^2 \right) \\ & \quad - \frac{1}{2}\exp\left(-2\left(x + \frac{\theta}{50} + \frac{3}{2}\right)^2\right)  - \frac{5}{7}\exp(-3x^2) \\ & \quad + \frac{1}{2}\exp\left(-4\left(x+\frac{3}{4}\right)^2\right)\\ & \quad + \frac{\theta}{5} \left[\frac{1}{2} \exp\left(-8\left(x+\frac{3}{2}\right)^2\right) + \frac{1}{2}\exp(-8x^2) +\exp\left(-8\left(x-\frac{3}{4}\right)^2\right) + \exp\left(-8\left(x+\frac{3}{4}\right)^2\right) \right]
% \end{align*}

% Here, we take $\mathcal{X} = [-2,2]$ and $\mathcal{S}_{\Theta} = \{-5,...,5 \}$ with $P(\theta = k) = \frac{|k|+1}{\sum_{j=-5}^5|j|+1}$.

\section{Derivation of closed-form kernels in Equation (13)}\label{app:closed_form_proof}

We show below a detailed derivation of the closed-form functions $s_0^2(\cdot,\cdot)$ and $h(\cdot,\cdot)$, which facilitate the closed-form surrogate modeling of the objective $g(\cdot)$.

Consider first the kernel $s_0^2(\mathbf{x},\mathbf{x}') = \text{Cov}\{g(\mathbf{x}),g(\mathbf{x})\}$, and define:
\[k_{\mathcal{X}}(\mathbf{x},\mathbf{x}')= \sigma^2\exp \left\{-\sum_{j=1}^d\frac{(x_j-x_j')^2}{2\ell_j^2}\right\}.\]
Let $\phi(\cdot)$ be the standard normal density function. Following \cite{frazier-TP-2022}, this kernel can be represented via the integral form:
\begin{align}
    s_0^2(\mathbf{x},\mathbf{x}') &= \int_{\mathbf{z}}\int_{\mathbf{z}'}K[(\mathbf{x},\mathbf{z}),(\mathbf{x}',\mathbf{z}')] dP(\mathbf{z}) dP(\mathbf{z}')\\
    &=\int_{\mathbf{z}}\int_{\mathbf{z}'}K_\mathcal{X}(\mathbf{x},\mathbf{x}') \exp\left(-\frac{1}{2}\left\{ \sum_{l=1}^q\frac{(z_l-z_l')^2}{\gamma_l^2}\right\} \right)\phi(\mathbf{z}')\phi(\mathbf{z})d\mathbf{z} d\mathbf{z}'\\
    &=K_\mathcal{X}(\mathbf{x},\mathbf{x}')\int_{\mathbf{z}}\int_{\mathbf{z}'}\exp\left(-\frac{1}{2}\sum_{l=1}^q\frac{(z_l-z_l')^2}{\gamma_l^2} \right)\phi(\mathbf{z})\phi(\mathbf{z}')d\mathbf{z} d\mathbf{z}'\\
    &=K_\mathcal{X}(\mathbf{x},\mathbf{x}') \prod_{l=1}^q\int_{z_l} \phi(z_l)\int_{z_l'}\exp\left(-\frac{(z_l-z_l')^2}{2\gamma_l^2} \right)\phi(z_l')dz_l' dz_l\\
    &=K_\mathcal{X}(\mathbf{x},\mathbf{x}') \prod_{l=1}^q\int_{z_l} \phi(z_l)\int_{z_l'}(2\pi)^{-1/2}\exp\left(-\frac{1}{2}\left[\frac{z_l^2}{\gamma_l^2}+(1+\gamma_l^{-2}){z_l'}^2-2\frac{z_lz_l'}{\gamma_l^2} \right]\right)dz_l' dz_l\\
    &=K_\mathcal{X}(\mathbf{x},\mathbf{x}') \prod_{l=1}^q\int_{z_l} \phi(z_l)\exp\left(-\frac{1}{2}\left[\gamma_l^{-2}-(\gamma_l^2+\gamma_l^4)^{-1} \right]z_l^2 \right)(\gamma_l^{-2}+1)^{-1/2}dz_l\\
&=K_\mathcal{X}(\mathbf{x},\mathbf{x}') \prod_{l=1}^q (2\pi)^{-1/2}(\gamma_l^{-2}+1)^{-1/2}\int_{z_l} \exp\left(-\frac{1}{2}\left[1+\gamma_l^{-2}-(\gamma_l^2+\gamma_l^4)^{-1} \right]z_l^2 \right)dz_l\\
    &=K_\mathcal{X}(\mathbf{x},\mathbf{x}') \prod_{l=1}^q \left[(\gamma_l^{-2}+1)\left(\gamma_l^{-2}+1 - (\gamma_l^2 +\gamma_l^4)^{-1}\right) \right]^{-1/2},
\end{align}
which completes the derivation.

The closed-form expression for function $h$ can be derived via the same argument as above, with $\phi(\mathbf{z})$ omitted from the integrand and the outer integration not performed.

\section{Proof of Proposition 1}

We begin from the definition of the $k$-TVR acquisition function. For a given $i = 1, \cdots, k$, define the event $\mathcal{A}_i = \left\{g(\mathbf{x}_{(i)})> g(\mathbf{x}_{(j)}), \;\forall j\not=i\right\}\cap \left\{g(\mathbf{x}_{(i)})>g(\mathbf{x}_n^*) \right\} $. Note that:
\begin{align}
    \textup{$k$-TVR}(\mathcal{D}_{(k)}) &= \mathbb{E}\left[\mathds{1}_{\left\{\max_{i = 1, \cdots, k} g(\mathbf{x}_{(i)})>g(\mathbf{x}_n^*) \right\}}\text{VR}^{(k)}_n\left(\arg\max_{i = 1, \cdots, k} g(\mathbf{x}_{(i)});\mathcal{D}_{(k)}\right)\right]\\
    % \begin{split}
        &=\sum_{i=1}^k\mathbb{E}\left[\text{VR}^{(k)}_n\left(\arg\max_{i = 1, \cdots, k} g(\mathbf{x}_{(i)});\mathcal{D}_{(k)}\right) \Big| \mathcal{A}_i \right] \mathbb{P}\left(\mathcal{A}_i \right)\\
        &=\sum_{i=1}^k\text{VR}^{(k)}_n\left(\mathbf{x}_{(i)};\mathcal{D}_{(k)}\right) \mathbb{P}\left(\mathcal{A}_i \right).
    % \end{split}
    \end{align}
\noindent By Proposition 1 from \cite{azimi_batch2010}, it follows that $\mathbb{P}(\mathcal{A}_i) = p_{i,n}$, which completes the proof.

%     \begin{align}
%     \begin{split}
%         &=\sum_{i=1}^k\text{VR}^{(k)}_n\left(\mathbf{x}_{(i)},\mathcal{D}_{(k)}\right)P\left(\left\{g(\mathbf{x}_{(i)})> g(\mathbf{x}_{(j)}), \;\forall j\not=i\right\}\cap \left\{g(\mathbf{x}_{(i)})>g(\mathbf{x}_n^*) \right\} \right)
%     \end{split}\\
%     \begin{split}
%         &\stackrel{\text{def}}{=}\sum_{i=1}^k\text{VR}^{(k)}_n\left(\mathbf{x}_{(i)},\mathcal{D}_{(k)}\right)p_{i,n}
%     \end{split}
% \end{align}

\section{Continuity extension of the $k$-TVR}
We provide here further details on the continuity extension of the batched TVR, following analogously from the continuity extension discussed in Section 4.1. Note that, in Equation (21), the $k$-TVR acquisition is defined only when there are no duplicates in the point set $\{\mathbf{x}_{(1)}, \cdots, \mathbf{x}_{(k)}, \mathbf{x}_n^*\}$. We outline below a slight and natural continuity extension, which holds for all choices of $\{\mathbf{x}_{(1)}, \cdots, \mathbf{x}_{(k)}\}$ for batch acquisition optimization.

For a given $\mathcal{D}_{(k)}=\{(\mathbf{x}_{(1)},\boldsymbol{\theta}_{(1)}), \cdots,(\mathbf{x}_{(k)},\boldsymbol{\theta}_{(k)}) \}$, define $\mathcal{D}_{(k)}^*$ as the unique points in $\{\mathbf{x}_{(1)}, \cdots, \mathbf{x}_{(k)}\}$ and let $\tilde{\mathcal{D}}_k^* = \mathcal{D}_{(k)}^*\cup\{\mathbf{x}_n^*\}$. We then define the slightly-extended $k$-TVR acquisition function as:
\begin{equation}
    \textup{$k$-TVR}'(\mathcal{D}_{(k)}) = \sum_{\mathbf{x}\in \mathcal{D}_{(k)}^*}p'_\mathbf{x}\text{VR}^{(k)}_n\left(\mathbf{x};\mathcal{D}_{(k)}\right),
\end{equation}
where:
\begin{equation}
    p'_\mathbf{x} = \begin{cases}
        \prod_{\tilde{\mathbf{x}}\in \tilde{\mathcal{D}}_k^*\backslash \{\mathbf{x}\}}(1 - \Phi(-\alpha'(\tilde{\mathbf{x}}))),&\mathbf{x}\not=\mathbf{x}_n^*\\
        \frac{1}{2}\prod_{\tilde{\mathbf{x}}\in \mathcal{D}_{(k)}^*\backslash \{\mathbf{x}\}}(1 - \Phi(-\alpha'(\tilde{\mathbf{x}}))),&\mathbf{x}=\mathbf{x}_n^*.
    \end{cases}
\end{equation}
\noindent Here, we have $\alpha'(\mathbf{x}) = \left(\mathbf{A}_{\mathbf{x}}\boldsymbol{\Sigma}_{n,k^*} \mathbf{A}_{\mathbf{x}}^\top \right)^{-1/2} \mathbf{A}_\mathbf{x} \boldsymbol{\mu}_{n,k^*}$, where $\mathbf{A}_\mathbf{x}$, $\boldsymbol{\Sigma}_{n,k^*}$ and $\boldsymbol{\mu}_{n,k^*}$ are defined analogously as in Proposition 1 using the elements in $\tilde{\mathcal{D}}_k^*$.

\section{Experiments with existing two-stage methods}

As mentioned in the discussion of benchmarks in Section 5, we employed a minor modification of the two-stage approach in \cite{Williams2000} for our experiments. For the first stage, we used $\mu_n(\mathbf{x}_n^*)$ as the current best solution in the EI formulation rather than an expensive Monte Carlo approximation of the acquisition. This facilitates a quicker (and more effective) EI optimization given a comparable computational budget for acquisition optimization with other methods.

 % The difference with \cite{Williams2000} is to use $\mu_n(\mathbf{x}_n^*)$ as the current ``best" solution used by the EI portion (rather than using Monte carlo approximations of EI with only the previously queried $\mathbf{x}$ points as possible ``best" points) in order to mirror the way that terminal solutions are selected in our experiments. Additionally, the use of VR in the second step is very similar to the look-ahead integrated mean-squared-prediction-error criterion used by \cite{Williams2000} when $\mathbf{x}_{n+1}$ is fixed, with simpler interpretation and computation.

\spacingset{1.0}
\bibliography{TVR_references}